\documentclass[dvipsnames,format=sigconf,anonymous=false,review=false,nonacm=true]{acmart}
\AtBeginDocument{%
  }

\setcopyright{acmlicensed}
\copyrightyear{2018}
\acmYear{2018}
\acmDOI{XXXXXXX.XXXXXXX}
\acmISBN{978-1-4503-XXXX-X/18/06}

\usepackage{amsthm}
\usepackage{graphicx}
\usepackage{cancel} 
\usepackage{amsmath}
\usepackage{multirow}
\usepackage{tabularx}
\usepackage{algorithm}
\usepackage{algorithmic}

\newtheorem{myProp}{Prop}
\begin{document}

\title{A Novel Two-Phase Cooperative Co-evolution Framework  for \\ Large-Scale Global Optimization with Complex Overlapping}

\author{Wenjie Qiu}
\email{wukongqwj@gmail.com}
\orcid{0009-0000-1965-8863}
\affiliation{%
  \institution{South China University of Technology}
  \city{Guangzhou}
  \state{Guangdong}
  \country{China}
}

\author{Hongshu Guo}
\email{guohongshu369@gmail.com}
\orcid{0000-0001-8063-8984}
\affiliation{%
  \institution{South China University of Technology}
  \city{Guangzhou}
  \state{Guangdong}
  \country{China}
}

\author{Zeyuan Ma}
\email{scut.crazynicolas@gmail.com}
\orcid{0000-0001-6216-9379}
\affiliation{%
  \institution{South China University of Technology}
  \city{Guangzhou}
  \state{Guangdong}
  \country{China}
}

\author{Yue-Jiao Gong}
\email{gongyuejiao@gmail.com}
\authornote{Corresponding Author}
\orcid{0000-0002-5648-1160}
\affiliation{%
  \institution{South China University of Technology}
  \city{Guangzhou}
  \state{Guangdong}
  \country{China}
}

\renewcommand{\shortauthors}{Qiu et al.}

\begin{abstract}
Cooperative Co-evolution, through the decomposition of the problem space, is a primary approach for solving large-scale global optimization problems. Typically, when the subspaces are disjoint, the algorithms demonstrate significantly both effectiveness and efficiency compared to non-decomposition algorithms. However, the presence of overlapping variables complicates the decomposition process and adversely affects the performance of cooperative co-evolution. In this study, we propose a novel two-phase cooperative co-evolution framework to address large-scale global optimization problems with complex overlapping. An effective method for decomposing overlapping problems, grounded in their mathematical properties, is embedded within the framework. Additionally, a customizable benchmark for overlapping problems is introduced to extend existing benchmarks and facilitate experimentation. Extensive experiments demonstrate that the algorithm instantiated within our framework significantly outperforms existing algorithms. The results reveal the characteristics of overlapping problems and highlight the differing strengths of cooperative co-evolution and non-decomposition algorithms. Our work is open-source and accessible at: https://github.com/GMC-DRL/HCC.
\end{abstract}



\maketitle

\section{Introduction}
\label{intro}
Large-Scale Global Optimization (LSGO) is one of most challenging problems in Evolutionary Computation (EC) \cite{jia2019distributed,ma2018survey,jian2020large}. A primary difficulty is the "curse of dimensionality": when the dimensionality of the problem space $\Omega$ becomes larger, the complexity of the problem and the resource overhead of algorithms increase exponentially \cite{jia2020contribution}. To address this, Cooperative Co-evolution (CC) employs a divide-and-conquer approach, decomposing the problem space into smaller subspaces $\Omega_i$ for optimization ($\Omega = \Omega_1 \times \dots \times \Omega_m$) \cite{ma2018survey}. Another major difficulty is the unknown problem structure: LSGO is also a type of Black-Box Optimization, the objective function is either unknown or too complex to be explicitly defined \cite{ma2024metabox,lian2024rlemmo}. 
This lack of a clear problem structure prevents it from being used to effectively guide the decomposition of the problem space \cite{ma2024toward,ma2024neural}. 
The ideal decomposition of the $D$-dimensional optimization problem $f(\mathbf{x}):\mathbb{R}^n\rightarrow\mathbb{R}$ \cite{omidvar2013cooperative}:
\begin{equation}
\label{ideal decomposition}
    \underset{\mathbf{x}_{1}, \ldots, \mathbf{x}_{m}}{\arg \min } f(\mathbf{x})=\left(\underset{\mathbf{x}_{1}}{\arg \min } f\left(\mathbf{x}_{1}, \ldots\right), \ldots, \underset{\mathbf{x}_{m}}{\arg \min } f\left(\ldots, \mathbf{x}_{m}\right)\right)
\end{equation}
That is, the optimal solution of the problem space is the union of the optimal solutions of the subspaces, where the subspaces $\Omega_i$ are disjoint. In other words, there is no interaction between them \cite{sun2015extended}.
Many studies have highlighted the importance of decomposition accuracy and have proposed various effective strategies to identify variable interactions and decompose the problem space when the structure is unknown, such as DG2 \cite{omidvar2017dg2}, RDG3 \cite{sun2019decomposition}, GDG \cite{mei2016competitive} and CSG \cite{tian2024composite}, among others \cite{chen2010large,omidvar2010cooperative,hu2017cooperation,sun2017recursive,yang2020efficient,zhang2019dynamic}.

However, problems involving interactions among different subspaces are prevalent in real-world applications \cite{wen2016maximal}, such as multi-silo problems \cite{ibrahimov2012evolutionary} in supply chain optimization and large-scale virtual network embedding problems \cite{song2019divide}. Variables that belong to two or more subspaces are referred to as overlapping variables, and the corresponding problem are called overlapping problem \cite{komarnicki2024overlapping}. The overlapping problem
adds extra complexity to LSGO. \cite{tian2024enhanced}. 
A series of large-scale variants of EC algorithms, known as non-decomposition algorithms (NDAs), do not encounter the complexity of decomposition and have gradually demonstrated superior performance across various problems compared to CC. \cite{akimoto2016projection,loshchilov2018large,loshchilov2017lm,ros2008simple,he2020mmes,li2017simple}.
How can the optimization performance of CC be further enhanced when dealing with complex overlapping problems? 
The complexity of overlapping problems for CC arises from both decomposition and optimization aspects. In terms of decomposition, achieving ideal decomposition remains a challenge for DG2 and other strategies in the presence of overlapping variables. But before asking how to achieve ideal decomposition, there is a fundamental question: Is ideal decomposition still important in this context? In terms of optimization, when the subspaces have no interaction, independent optimization is possible, but the question arises: how should they be optimized now? Moreover, given the competitiveness of NDAs, the question of when to use CC becomes crucial. Can CC leverages the advantages of NDAs to enhance its performance? To answer these questions and effectively improve CC's performance, it is essential to conduct a detailed investigation into the impact of overlap on both NDAs and CC.
However, the limited number of overlapping problems in existing benchmarks adds a difficulty to analyzing and addressing these questions \cite{xu2023large}.
Therefore, we make the following main contributions in this study:
\begin{itemize}
    \item A novel benchmark for overlapping problems called the Auto Overlapping Benchmark (AOB) is proposed to extend existing benchmarks. AOB allows researchers to easily define and automatically generate the required overlapping problems for analysis.
    \item A novel two-phase CC framework, called the Hybrid Cooperative Co-evolution Framework (HCC), is proposed to solve complex overlapping problems. This framework is developed based on an analysis of the optimization behavior of NDAs and CC on both non-overlapping and overlapping problems.
    \item A novel decomposition strategy, called Recursive Decomposition of Design Structure Matrix (RDDSM), is proposed and embedded within HCC. Following the first principle, RDDSM accurately identifies overlapping variables, thereby ensuring the effectiveness of decomposition in overlapping problems.
    \item  Extensive experiments demonstrate that the algorithm instantiated within HCC significantly outperforms existing algorithms. The results also highlight the differing strengths of NDAs and CC on overlapping problems, shedding light on the characteristics of overlapping problems and providing insights to address the aforementioned questions. 
\end{itemize}

The rest of this paper proceeds as follows. Section \ref{section:Background} provides a brief background on the overlapping problem. Section \ref{section:HCC} presents the detailed implementation of HCC. Section \ref{section:exp} introduces AOB, accompanied by extensive experiments that validate the superior performance of HCC and reveal some interesting insights. Section \ref{section:con} concludes with a summary and outlook.

\section{Background}
\label{section:Background}
In this section, we briefly describe the variable interaction and overlapping variable.
Let $\Omega_i, \Omega_j$
be two disjoint subspaces. They do not interact if conditions
\begin{equation}
\begin{aligned}
f(\mathbf{x}) &< f\left(\mathbf{x} + \delta_{i} \mathbf{u}_{i}\right) \Leftrightarrow f\left(\mathbf{x} + \delta_{j} \mathbf{u}_{j}\right) < f\left(\mathbf{x} + \delta_{i} \mathbf{u}_{i} + \delta_{j} \mathbf{u}_{j}\right) \\
f(\mathbf{x}) &= f\left(\mathbf{x} + \delta_{i} \mathbf{u}_{i}\right) \Leftrightarrow f\left(\mathbf{x} + \delta_{j} \mathbf{u}_{j}\right) = f\left(\mathbf{x} + \delta_{i} \mathbf{u}_{i} + \delta_{j} \mathbf{u}_{j}\right)
\end{aligned}
\end{equation}
hold for any vector $\mathbf{x} \in \Omega$, values $\delta_i$, $\delta_j > 0$, and unit vectors $\mathbf{u_i} \in \Omega_i, \mathbf{u_j} \in \Omega_j$ \cite{komarnicki2024overlapping,sun2015extended}. If $\Omega_i$ and $\Omega_j$ interact, this is denoted by $\Omega_i \leftrightarrow \Omega_j$, else $\Omega_i \nleftrightarrow \Omega_j$. Naturally, when $\Omega_i$ and $\Omega_j$ are one-dimension spaces, this interaction degenerates into an interaction between two variables $x_i$ and $x_j$, which is denoted by $x_i \leftrightarrow x_j$. For equation \ref{ideal decomposition}, ideal decomposition is equivalent to: 
\begin{equation}
\label{ideal decomposition 2.1}
    \forall \Omega_i, \Omega_j, \Omega_i \nleftrightarrow \Omega_j
\end{equation}
\begin{equation}
\label{ideal decomposition 2.2}
\forall \Omega_i, \forall x_p, x_q \in \Omega_i, x_p \leftrightarrow x_q
\end{equation}
That is, the subspaces are non-overlapping, but within each subspace, all variables are pairwise interacting.

\begin{figure}[t]
    \centering
    \includegraphics[width=1\linewidth]{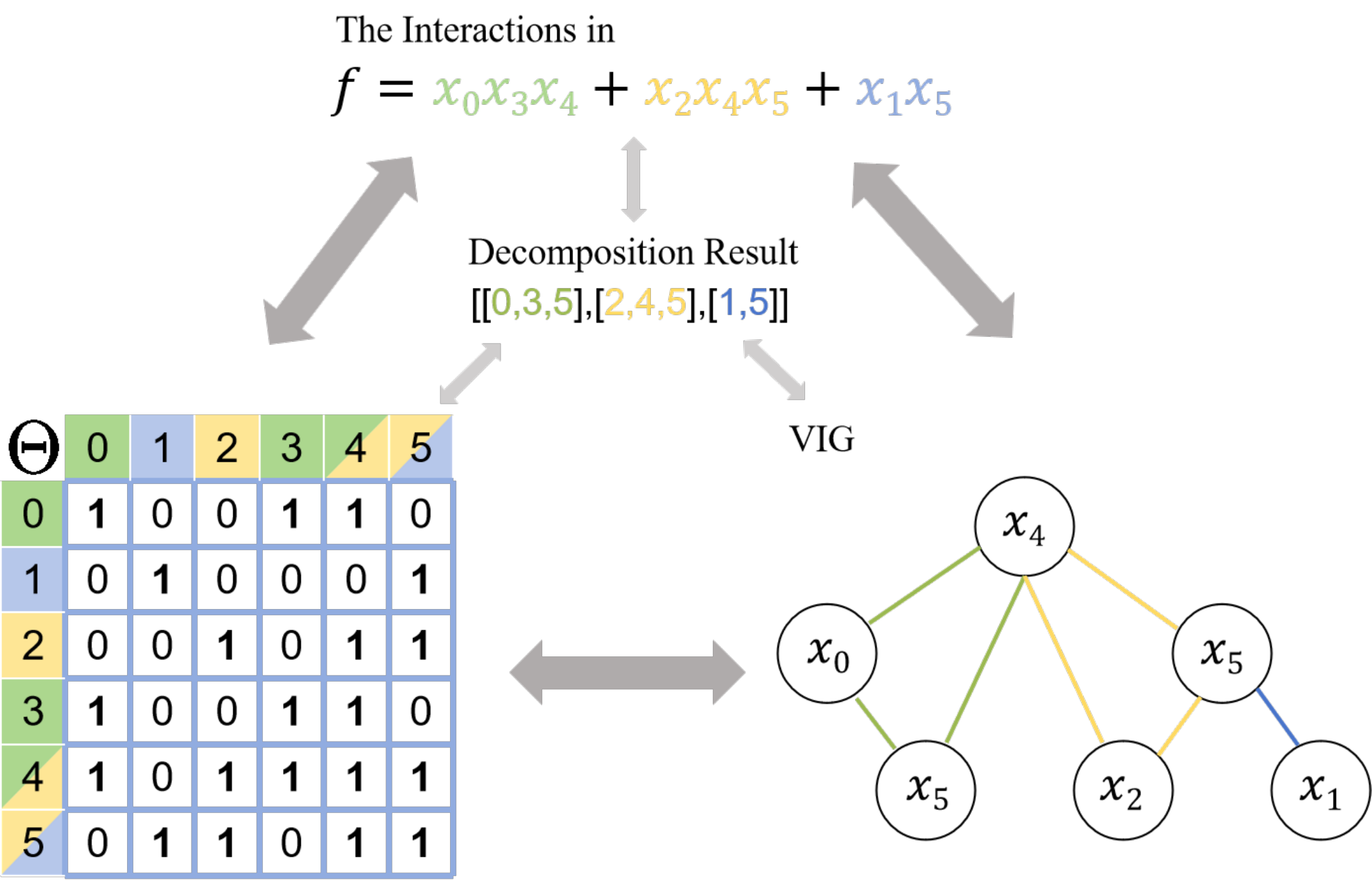}
    \caption{The equivalence of representations of interactions.}
    \label{Equivalence}
    \vspace{-0.6cm}
\end{figure}

Although the problem structure is unknown, many decomposition strategies can identify interactions between variables. These strategies classify interactions into various types, such as additive separable \cite{tang2007benchmark}, multiplicative separable \cite{li2022dual}, and composite separable \cite{tian2024composite}. When the problem space consists of additively separable subspaces, a decomposition strategy called Differential Grouping (DG) is effective \cite{omidvar2013cooperative}, particularly its variant DG2 \cite{omidvar2017dg2}. 
DG2 outputs the design structure matrix $\Theta$, which records the interaction information among all variables. The elements value $\theta_{p,q}$ of $\Theta$, which are either 1 or 0, indicate whether $x_p$ and $x_q$ interact. $\Theta$ is equivalent to the variable interaction graph (VIG), where nodes represent variables and edges represent interactions. Figure \ref{Equivalence} illustrates a specific example of this equivalence. 
The overlapping variables, such as $x_4, x_5$, interact with variables in two or more subspaces. This naturally breaks the condition that  $\Omega_i, \Omega_j$ are two disjoint subspaces, making it impossible to achieve equation \ref{ideal decomposition 2.1}. Therefore, for overlapping problems, the ideal decomposition is achieved by satisfying equation \ref{ideal decomposition 2.2} while ensuring that interactions caused by overlapping variables are excluded, ultimately satisfying equation \ref{ideal decomposition 2.1}. Many studies have extended this idea to address the overlapping problem, such as RDG3, DOV \cite{meselhi2022decomposition}. 
However, the decompositions obtained are often tailored to specific problems, and there is a lack of mathematical proof for the decomposition to achieve ideal decomposition. Therefore, We propose RDDSM, a method that is mathematically proven to achieve ideal decomposition of overlapping problems.


The benchmark for the CEC’2013 Special Session and Competition on Large-Scale Global Optimization (CEC2013LSGO) is one of the most commonly used in LSGO \cite{li2013benchmark}. CEC2013LSGO consists of 15 problems, but only two of them involve overlapping variables. The analysis of overlapping problems is constrained by the limited availability of such problems in existing benchmarks. In an LSGO benchmark, several components are key to constructing a problem $F$: 1) Base function $f$ 2) The list of subspaces size $\mathcal{S}_{size}$ 3) The random permutation of the dimension indices $\mathcal{P}$  4) The shift vector $\mathbf{x}^{opt}$ 5) The list of subspaces weight $w$ 6) The list of subspaces orthogonal rotation matrix of $\mathcal{R}$ 7) The list of number of overlapping variable in each subspace $\Gamma$. They are also fixed and researchers cannot customize these components to control various factors, adding more unknown factors that influence the analysis of overlap. Therefore, we propose AOB to extend CEC2013LSGO, enabling researchers to easily define components and automatically generate the required overlapping problems for analysis. We instantiated 24 problems in AOB (Section \ref{subsection:AOB}), covering a range from non-overlapping cases to scenarios where each subspace has up to 10 overlapping variables at varying degrees. These problems include multiple base functions such as Schwefel and Elliptic.

\section{Hybrid Cooperate Co-evolution
Framework}
\label{section:HCC}
\subsection{The Analysis of NDAs and CC Performance on CEC2013LSGO}
\label{section:anl of CEC}
F11 and F13 are two problems in CEC2013LSGO that both use Schwefel \cite{potter1994cooperative} as the base function. The main difference is that F11 is non-overlapping, while F13 is overlapping. Therefore, we conduct a simple comparison of NDAs and CC on these two problems to analyze their optimization performance and characteristics on non-overlapping and overlapping problems. CC include Random-CMAES, RDG3-CMAES, DG2-CMAES and RDDSM-CMAES, representing decomposition accuracies from low to high. "Random" refers to the random decomposition of 1000 dimensions into 20 subspaces. The settings for the decomposition strategies follow the configurations in their original papers \cite{duan2024pypop7,huang2024evox}. CMAES is an evolutionary algorithm used as the optimizer for subspaces within the CC \cite{auger2012tutorial}. The initial mean $\omega$ is a zero vector, the initial step size $\sigma$ is 0.5, and the population size $P$ is $4+3*\lceil log(|\Omega_i|) \rceil$ \cite{jia2019distributed}. The function evaluations (FEs) of decomposition are not considered and the total FEs (TFEs) is set to 3E6. The NDAs include Sep-CMAES \cite{ros2008simple}, LM-MA-ES \cite{loshchilov2018large}, LMCMA \cite{loshchilov2017lm}, MM-ES \cite{he2020mmes}, also with the consistent configurations in their original papers \cite{duan2024pypop7,huang2024evox}. Each experimental result is the average of 25 independent runs.

\begin{figure}[t]
    \centering
    \includegraphics[width=1\linewidth]{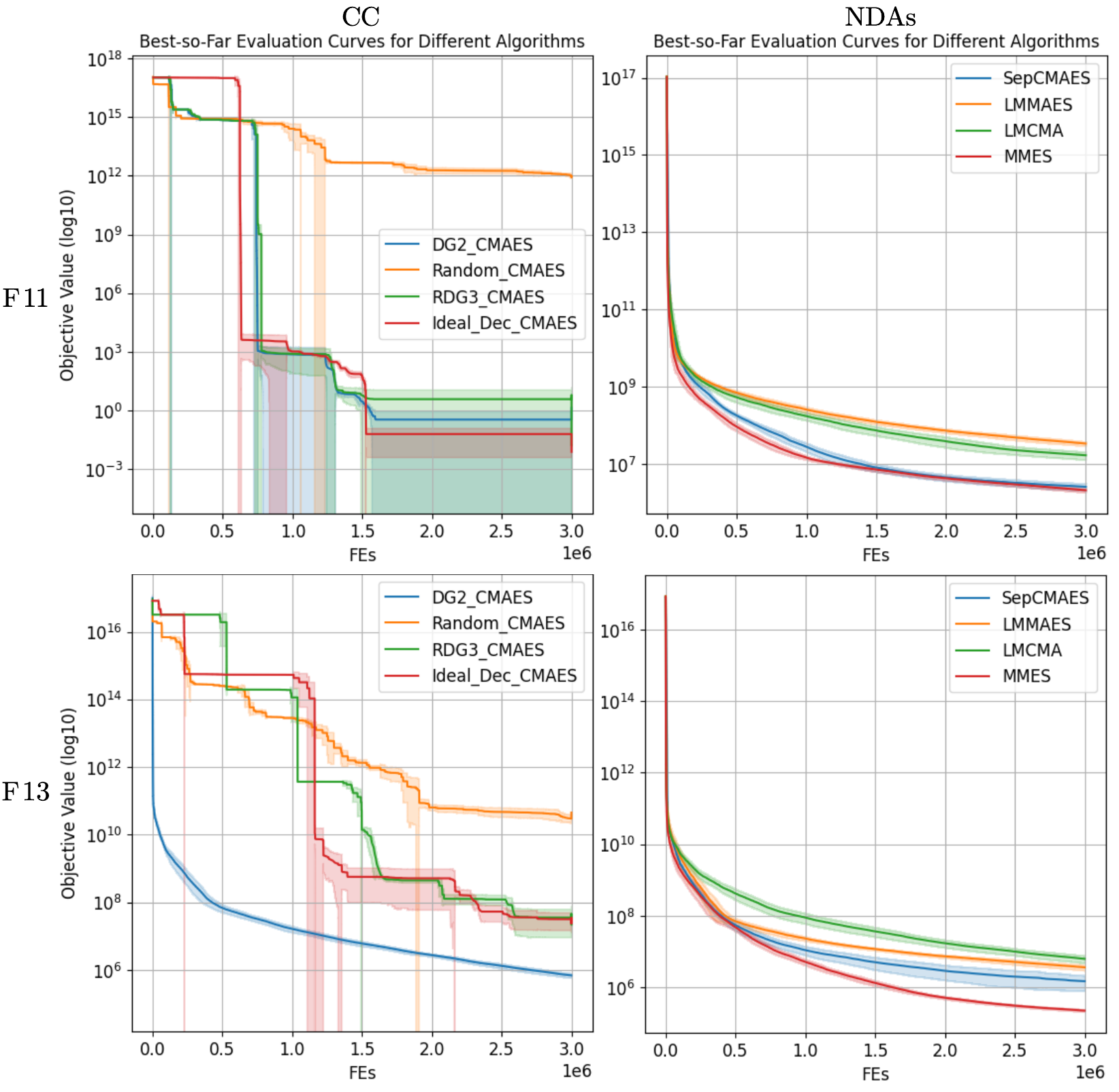}
    \caption{The Best-so-Far Evaluation Curves for Different Algorithms}
    \label{fig:best_so_far}
    \vspace{-0.6cm}
\end{figure}

\begin{figure}[t]
    \centering
    \includegraphics[width=1\linewidth]{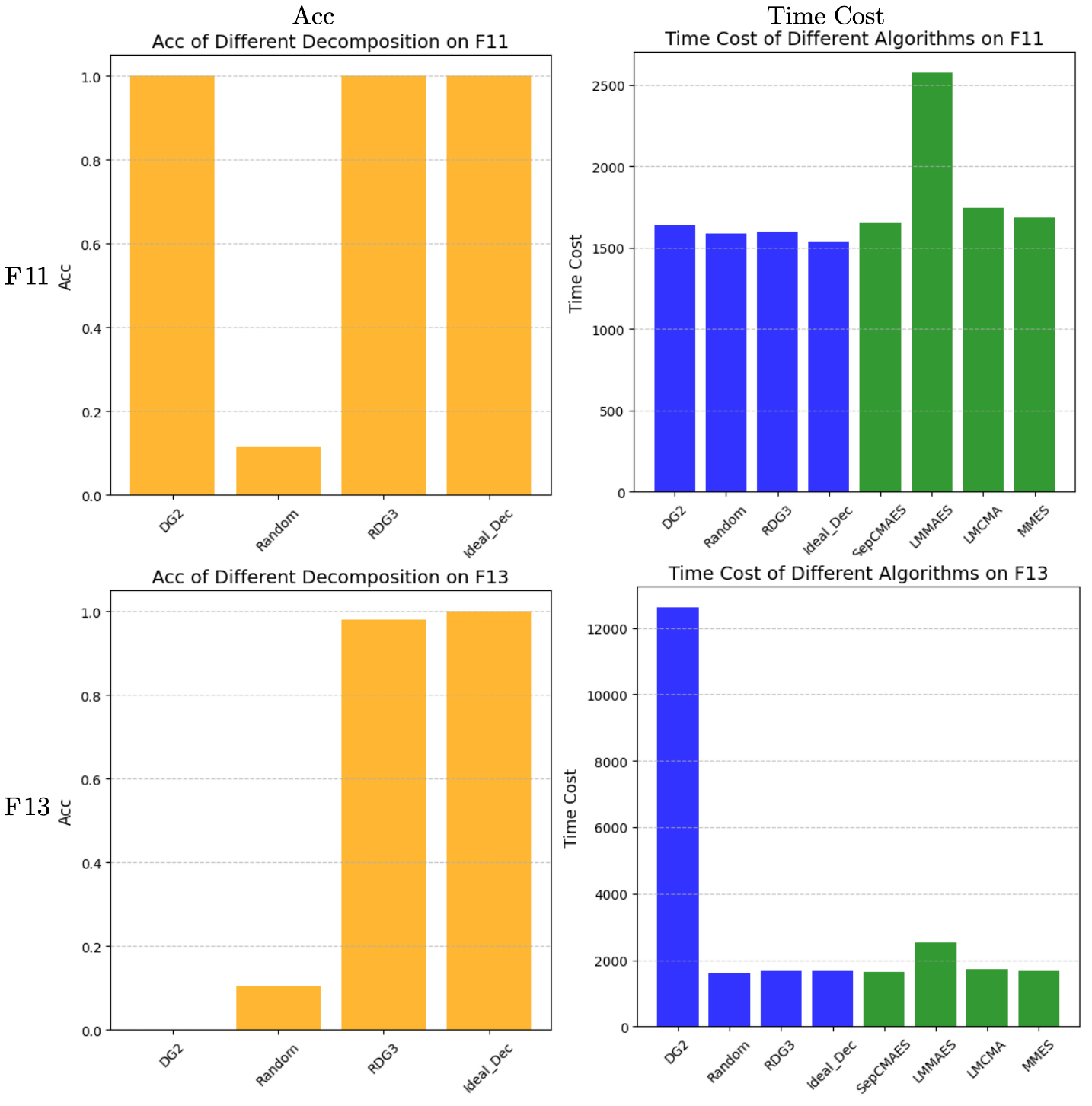}
    \caption{The Accuracy of Decomposition strategies and the Time Cost of Algorithms}
    \label{fig:resource_CEC}
    \vspace{-0.3cm}
\end{figure}

Figure \ref{fig:best_so_far} shows the best-so-far evaluation curves of different algorithms on F11 and F13. In terms of optimization performance, CC significantly outperforms NDAs on F11, but the opposite is true for F13. This demonstrates that overlap does indeed impact the effectiveness of CC. The potential reason for this impact stems from the optimization of the same variable across multiple subspaces, leading to different values. This is referred to as the coupling issue of the optimized values of overlapping variables. In terms of optimization behavior, the objective value for CC mainly decreases during the middle and later stages, often exhibiting a "staircase-like" pattern. In contrast, NDAs show a significant decrease in the early stages, with the middle and later stages tending to stabilize. This is likely because, in the early stages, CC has not yet identified better global optimization points, and the optimization effect relying solely on subspace optimization is limited. In Figure \ref{fig:best_so_far}, two notable phenomena can be observed: first, on F13, DG2-CMAES exhibits optimization behavior similar to that of NDAs and achieves exceptionally good performance. Second, CC shows greater variance, while NDAs exhibit smaller variance and are relatively more stable.

To analyze the first point, we calculated the decomposition accuracy of various decomposition strategies in CC and the time cost of all algorithms. The accuracy of decomposition (Acc) is calculated using the following equation:
\begin{equation}
    \text{Acc}=\frac{\sum_{i}^{m} \max \left\{\left|\boldsymbol{\Omega}_{i}^{*} \cap \boldsymbol{\Omega}_{1}\right|,\left|\boldsymbol{\Omega}_{i}^{*} \cap \boldsymbol{\Omega}_{2}\right|, \ldots,\left|\boldsymbol{\Omega}_{i}^{*} \cap \boldsymbol{\Omega}_{k}\right|\right\}}{D}
\end{equation}
where each true subspace $\Omega_{i}^{*}$, we find the subspace $\Omega_{i}$ contains the highest number of common variables among all the subspaces in a $D$-dimensions problem space. Figure \ref{fig:resource_CEC} shows the comparison results. 
\begin{figure}[t]
    \centering
    \includegraphics[width=1\linewidth]{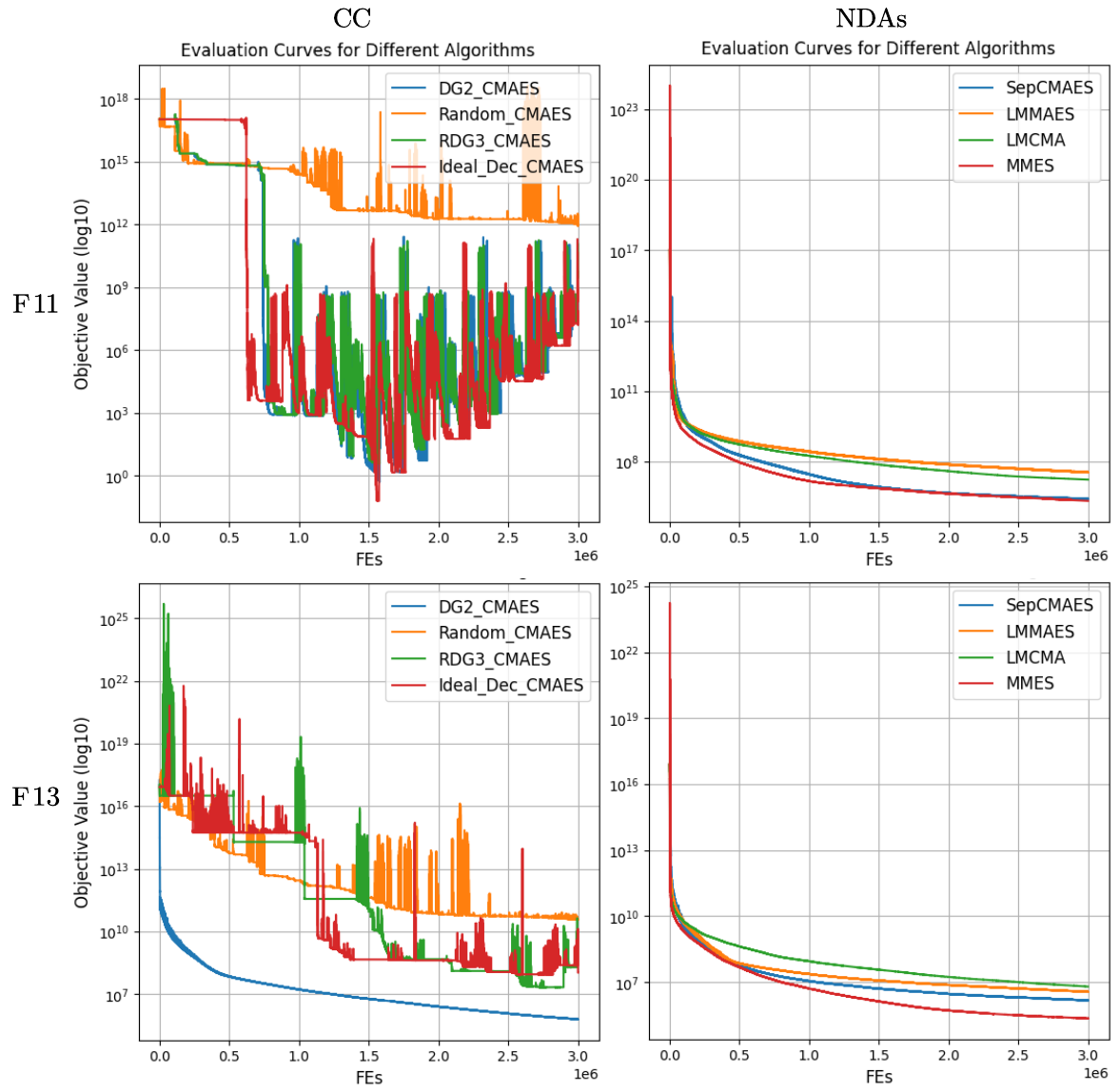}
    \caption{The Evaluation Curves for Different Algorithms}
    \label{fig:curve_CEC}
    \vspace{-0.6cm}
\end{figure}
A direct conclusion is that DG2 fails in decomposition on F13. In practice, CMAES operates as an NDA optimizer, which aligns with its observed optimization behavior. However, since CMAES is not specifically designed for LSGO, it cannot handle the "curse of dimensionality," resulting in significant time cost. This is because DG2 uses depth-first search to process the design structure matrix, and once it encounters overlapping problems, the decomposition fails. Excluding DG2's performance on F13, it can be observed that the more accurate the decomposition, the better the optimization performance. This holds true for both overlapping and non-overlapping problems. Therefore, achieving ideal decomposition is equally important for overlapping problems. 

To analyze the stability of CC and NDAs, we recorded the evaluation curves of all algorithms, and the results are presented in Figure \ref{fig:curve_CEC}.  
It is evident that CC exhibits significantly higher variability compared to NDAs, which is also reflected in the best-so-far curves. This may be due to two reasons: The primary reason is the coupling issue of the optimized values of overlapping variables. This is because directly altering the values of overlapping variables in one subspace can significantly affect other subspaces, rendering their optimization ineffective. This can be demonstrated in Figure \ref{fig:curve_CEC}, where the differences in CC's performance during the early stages on F11 and F13 are evident. F11 is non-overlapping, as evidenced by the minimal fluctuations observed in the early stage (before 5E5). The second reason is that when optimization stagnates, CC optimizes sequentially within smaller subspaces, leading to stronger exploratory behavior. This is reflected in its later middle and stages oscillating evaluation curve and "staircase-like" best-so-far evaluation curve. In contrast, NDAs face a much larger search space with an overwhelming number of exploration possibilities, leading them to focus more on exploitation along existing trajectories, resulting in relatively stable performance \cite{ma2024auto,guo2024deep}.

Based on the above analysis, the questions in Section \ref{intro} can be rephrased as the following:  
(1) How can ideal decomposition be achieved reliably and stably?  
(2) How can the coupling issue of the optimized values of overlapping variables be effectively addressed?  
(3) How can the exploitation strength of NDAs and the exploration strength of CC be balanced to overcome the slow objective value decline in the early stages of CC and the convergence in the later stages of NDAs?  

\subsection{Recursive Decomposition of Design Structure Matrix}
\label{section:RDDSM}
The decomposition strategies used by DG2 is depth-first search, which becomes ineffective for overlapping problems. To address this issue, RDG3 introduces a method to control the size of subspaces, but it fails to achieve ideal decomposition. Subsequently, many other strategies have utilized statistics-based or heuristic methods to decompose the design structure matrix \cite{wen2016maximal,jia2019distributed,meselhi2022decomposition,tian2024enhanced,zhang2019dynamic,omidvar2010cooperative}. However, these methods face one major challenge: they are designed for specific scenarios, which often makes it difficult to consistently achieve ideal decomposition when faced with different overlapping situations \cite{omidvar2021review1,omidvar2021review2}.

The above issues arise from not considering the inherent characteristics of the design structure matrix. Therefore, we propose the Recursive Decomposition of Design Structure Matrix (RDDSM).
To begin, it is necessary to introduce a series of basic concepts. $\Theta[i,:]$ is the $i-$th row vector of $\Theta$. Positive component vector $P_i$ is a vector composed of the dimensions of $\Theta[i,:]$ being 1. Principal submatrix $M_i$ is the submatrix formed by extracting the corresponding columns and rows of $\Theta$ from the each dimension of $P_i$. Then we prove two important properties in the $\Theta$.
\begin{myProp}
\label{prop1}
If $\Theta[i,:]$ is different from $\Theta[j,:]$, then the subspace sets of variables $i$ and $j$ are distinct.
\end{myProp}
Proof by contradiction: Assume variables $i$ and $j$ have the same subspace sets. If $\Theta[i,k]=1$, this implies that variables $i$ and $k$ are interactive in a subspace. Since
variables $i$ and $j$ are assumed to have the same subspace sets, variables $j$ and $k$ must also be interactive in the same subspace, which would imply $\Theta[j,k] = 1$. This contradicts our
assumption, completing the proof.
\begin{myProp}
\label{prop2}
$M_i$ is an all-1 matrix if and only if all variables in $M_i$ are interactive.
\end{myProp}
Proof by contradiction: Assume variables $p$ and $q$ are not interactive, then $M_i[p,q] = 0$;
Assume $M_i[p,q] = 0$, then the variables $p$ and $q$ are not interactive. These contradict our
assumption, completing the proof.

\begin{figure}[t]
    \centering
    \includegraphics[width=1\linewidth]{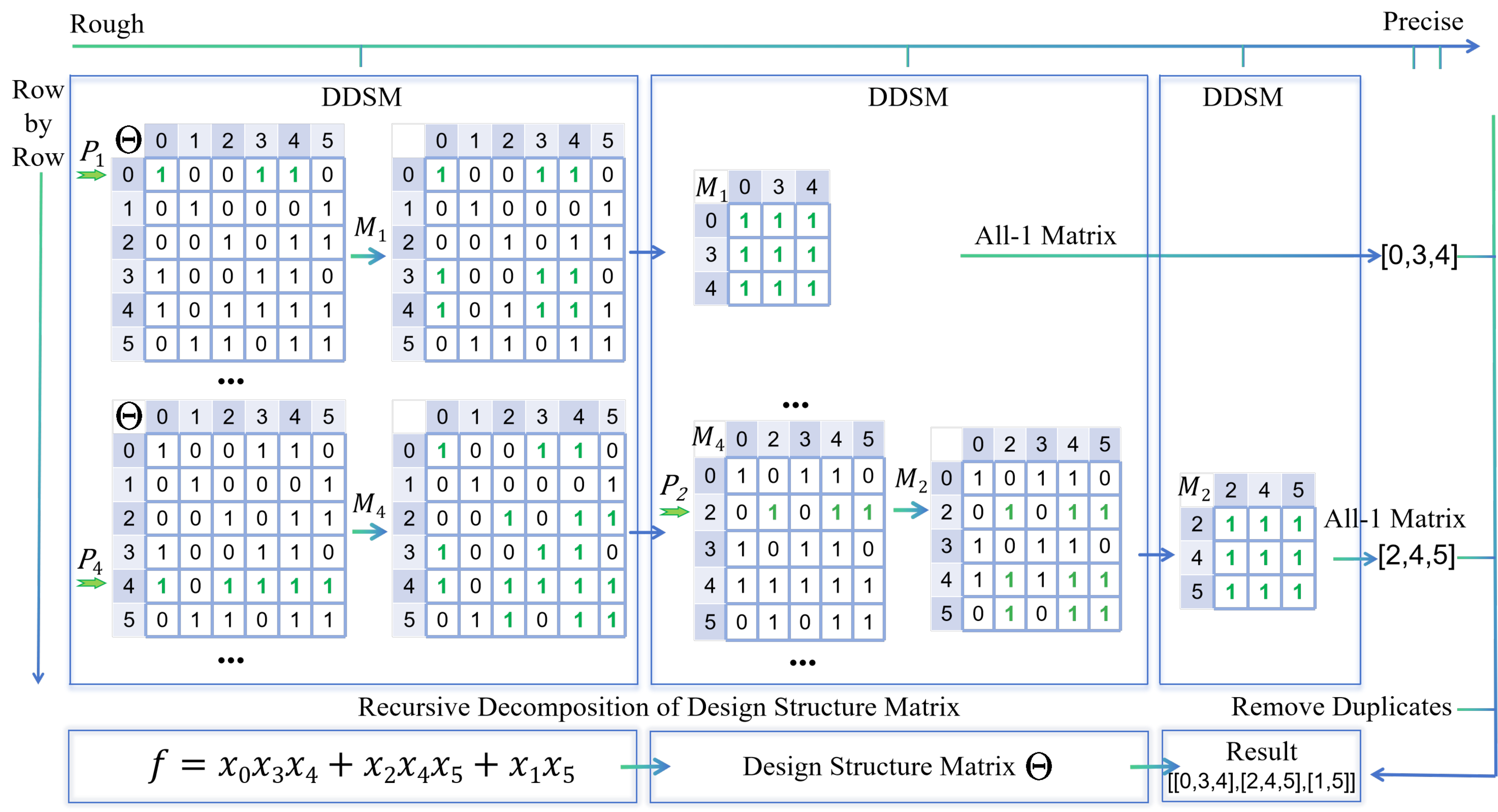}
    \caption{The Process of RDDSM}
    \label{RDDSM}
    \vspace{-0.4cm}
\end{figure}

It should be noted that $\Theta$ cannot provide certain interaction information. For example, $f_1=x_1x_2x_3+x_2x_3x_4+x_1x_4$ corresponds to all-1 $\Theta$, which is consistent with $f_2=x_1x_2x_3x_4$. However, $\Theta$ cannot provide any additional information for decomposing to $f_1$. Therefore, $\Theta$ has an upper bound for decomposition information. Considering that such overlapping situations can be very complex and varied, we uniformly assume that an all-1 matrix corresponds to interactions between variables within a subspace. Under this setting, although equation \ref{ideal decomposition 2.1} is unable to achieve in overlapping problems, equation \ref{ideal decomposition 2.2} is able to achieve. Figure \ref{RDDSM} illustrates the process of RDDSM and pseudocode \ref{RDDSM} shows the flow of RDDSM. According to property 2, we need to find the all-1 $M$ in $\Theta$. In RDDSM, we check the $\Theta$ row by row (DDSM's Line 3). If $M_i$ is an all-1 matrix, then, by property 2, all variables are grouped together and appended to the the list of Subspaces $\mathcal{S}$ as a single list (DDSM's Lines 4-5). If $M_i$ is a non-all-1 matrix, there exists an $\Theta[i,:]$ that differs from $\Theta[j,:]$, and by property 1 the subspace sets of variables $i$ and $j$ are distinct. The above process is called decomposition of design structure matrix (DDSM). It should be pointed out that M is relative. After completing above DDSM, non-all-1 $M$ is the new $\Theta$. Therefore, We can then recursively repeat the DDSM on non-all-1 $M_i$ and add the result to $\mathcal{S}$ (DDSM's Lines 7-8). Finally,remove any duplicate subspaces from $\mathcal{S}$ (DDSMs's Line 3). This whole process is called RDDSM.

\begin{algorithm}[t]
    \caption{RDDSM}
    \label{rddsm}
    \renewcommand{\algorithmicrequire}{\textbf{Input:}}
    \renewcommand{\algorithmicensure}{\textbf{Output:}}
    \begin{algorithmic}[1]
        \REQUIRE Design structure matrix $\Theta$
        \ENSURE The list of Subspaces $\mathcal{S}$ 
        \STATE Initialize $\mathcal{S}$ to be an empty list
        \STATE $\mathcal{S} \gets \text{DDSM} (\Theta, \mathcal{S})$ 
        \STATE $\mathcal{S} \gets \text{Set} (\mathcal{S})$
        \RETURN $\mathcal{S}$
    \end{algorithmic}
    \renewcommand{\algorithmicensure}{\textbf{Function:}}
    \hrule
    \begin{algorithmic}[1]
        \ENSURE DDSM ($\Theta$, $\mathcal{S}$)
        \STATE The length of matrix L $\gets \Theta$ 
        \WHILE{$i<=$ L}
            \STATE Get $M_i$ from $\Theta$
            \IF{$M_i$ is an all-1 Matrix}
                \STATE Add the list of variables in $M_i$ to $\mathcal{S}$ 
            \ELSE
                \STATE $\mathcal{S'} \gets$ DDSM ($M_i$, $\mathcal{S}$)
                \STATE $\mathcal{S} \gets \mathcal{S} \cup \mathcal{S'}$
            \ENDIF
        \ENDWHILE
        \RETURN $\mathcal{S}$
    \end{algorithmic}
\end{algorithm}

In theory, RDDSM can decompose the problem space into the most precise grouping under the upper bound of the decomposition information in $\Theta$. It should be noted that overlap is relative to the space. For example, in $f=x_0x_3x_4+x_2x_4x_5+x_1x_5$, $x_4$ is is considered an overlap variable with respect to the entire space. However, for the subspace corresponding to $x_0x_3x_4$, $x_4$ is non-overlapping. Therefore, transforming $\Theta$ into $M$ is essentially the process of dividing the space. And $M$ has only two types: all-1 and non-all-1. In 
$M_i$, the row vector corresponding to $i$ variable must be an all-1 row vector, because the definition of $M_i$. The all-1 case occurs only when variable 
$i$ is non-overlapping in the space corresponding to $M_i$. For the non-all-1 case, further subdivision is needed until an all-1 case appears. Therefore, RDDSM naturally divides the overlapping variables in the original space into its non-overlapping space. 

\begin{algorithm}[t]
    \caption{HCC}
    \label{HGCC}
    \renewcommand{\algorithmicrequire}{\textbf{Input:}}
    \renewcommand{\algorithmicensure}{\textbf{Output:}}
    \begin{algorithmic}[1]
        \REQUIRE Problem $F$, Design structure matrix $\Theta$, Total FEs $TFEs$
        \ENSURE The global best point $gbest$ 
        \STATE $\mathcal{S} \gets$ RDDSM ($\Theta$)
        \STATE Calculate $DO$ and $GloFEs$ based on Equation \ref{DO} and \ref{GloFEs}
        \STATE Initialize $gbest, \omega, \sigma, P, sumFEs$
        \STATE $gbest$, $\omega \gets \text{NDA} (gbest, F, \omega, \sigma, P, GloFEs)$
        \STATE $sumFEs \gets sumFES + GloFEs$ 
        \WHILE{the termination condition is not met} 
            \STATE $subFEs \gets \frac{(TFES-sumFEs)}{|\mathcal{S}|}$ 
            \FOR{$\mathcal{S}[i]$ in $\mathcal{S}$ }
                \STATE Initialize $\omega_i, \sigma_i, P_i$
                \STATE $gbest_i, \omega_i \gets gbest[\mathcal{S}[i]]$, $\omega[\mathcal{S}[i]] $ 
                \STATE $gbest_i, \omega_i, FEs \gets$ OPT ($gbest_i, F,\omega_i,\sigma_i,P_i,subFEs$) 
                \STATE Adjust $gbest_i, \omega_i$ based on Equation \ref{adjust}
                \STATE $sumFEs \gets sumFES + FEs$ 
            \ENDFOR
        \ENDWHILE 
        \RETURN $gbest$
    \end{algorithmic}
\end{algorithm}
\subsection{HCC}

\begin{table*}[t]
  \centering
  \caption{The Detail Configuration of AOB}
  \label{table:AOB}
  \vspace{-0.4cm}
  \resizebox{1\textwidth}{!}{
    \begin{tabular}{cccc}
    \toprule
    \multicolumn{4}{c}{Problem} \\
    \midrule
    \multicolumn{4}{c}{$F(\mathbf{z})=\sum_{j=1}^{|\mathcal{S}_{size}|} w_{i} f_{\text{Base}}\left(\mathbf{z}_{j}\right)$} \\
    \midrule
    \multicolumn{4}{c}{Base Functions Pool} \\
    \midrule
    \multicolumn{1}{c}{$f_{\text {Schwefel}}(\mathbf{x})=$} & \multicolumn{1}{c}{$\sum_{i=1}^{\mathcal{S}_{size}[j]}\left(\sum_{j=1}^{i} x_{i}\right)^{2}$} & \multicolumn{1}{c}{$f_{\text {Elliptic}}(\mathbf{x})=$} & \multicolumn{1}{c}{$\sum_{i=1}^{\mathcal{S}_{size}[j]} 10^{6 \frac{i-1}{\mathcal{S}_{size}[j]-1}} x_{i}^{2}$} \\
    \multicolumn{1}{c}{$f_{\text {Rastrigin}}(\mathbf{x})=$} & \multicolumn{1}{c}{$\sum_{i=1}^{\mathcal{S}_{size}[j]}\left[x_{i}^{2}-10 \cos \left(2 \pi x_{i}\right)+10\right]$} & \multicolumn{1}{c}{$f_{\text {Ackley}}(\mathbf{x})=$} & \multicolumn{1}{c}{$-20 \exp \left(-0.2 \sqrt{\frac{1}{\mathcal{S}_{size}[j]} \sum_{i=1}^{\mathcal{S}_{size}[j]} x_{i}^{2}}\right)-\exp \left(\frac{1}{\mathcal{S}_{size}[j]} \sum_{i=1}^{\mathcal{S}_{size}[j]} \cos \left(2 \pi x_{i}\right)\right)+20+e$} \\
    \midrule
    \multicolumn{4}{c}{The Configuration of Component Parts} \\
    \midrule
    \multicolumn{4}{c}{$\mathcal{S}_{size}=[ 50,50,25,25,100,100,25,25,50,25,100,25,100,50,25,25,25,100,50,25], \quad D = \sum_{i=1}^{|\mathcal{S}_{size}|} \mathcal{S}_{size}[i] = 1000, \quad  \mathbf{x} \in[-100,100]^{D}$} \\
    \multicolumn{4}{c}{$\mathcal{S}[i]=\mathcal{P}(\sum_{k=1}^{i-1} \mathcal{S}_{size}[k]:\sum_{k=1}^{i} \mathcal{S}_{size}[k]+\Gamma[i-1]), i\ne 1, \quad  \mathcal{S}[1]=\mathcal{P}(0:\mathcal{S}_{size}[1]), \quad 
 \mathcal{S}=[\mathcal{S}_1,\dots, \quad \mathcal{S}_{|\mathcal{S}_{size}|}]$} \\
    \multicolumn{4}{c}{$w_{i}=10^{3 \mathcal{N}(0,1)}, w = [w_1,\dots, \quad 
 w_{|\mathcal{S}_{size}|}], \quad \mathcal{R}_i=\text { a }\left|\mathcal{S}_{i}\right| \times\left|\mathcal{S}_{i}\right| \text { rotation matrix }, \quad \mathcal{R} = [\mathcal{R}_1,\dots,\mathcal{R}_{|\mathcal{S}_{size}|}]$} \\
    \multicolumn{4}{c}{$\mathrm{y}=\mathrm{x}-\mathrm{x}^{\mathrm{opt}} ,\quad 
 \mathbf{y}_{i}=\mathbf{y}[\mathcal{S}[i]], i \in\{1, \ldots,|\mathcal{S}|\} , \quad \mathbf{z}_{\mathbf{i}}=T_{\mathrm{asy}}^{0.2}\left(T_{\mathrm{osz}}\left(\mathcal{R}_{i} \mathbf{y}_{\mathbf{i}}\right)\right), i \in\{1, \ldots,|\mathcal{S}|\} $} \\
    \bottomrule
    \end{tabular}%
    }
  \label{tab:addlabel}%
\end{table*}%
RDDSM attempts to answer the first question posed in Section \ref{section:anl of CEC}: "How to stably achieve ideal decomposition?" But how should we address the remaining two questions? Based on the analysis in Section \ref{section:anl of CEC}, it is important to note that NDAs and CC are not mutually exclusive; rather, they can complement each other in solving LSGO problems. 
Therefore, a novel framework called Hybrid Cooperate Co-evolution Framework (HCC) is proposed. 

The core idea of HCC is to adjust the collaboration degree between NDAs and CC based on the degree of overlap ($DO$) to balance exploitation and exploration, and to address the coupling issue of the optimized values of overlapping variables by considering the contribution of each subspace. The degree of overlap $DO$ is calculated by the proportion of overlapping variables in a $D$-dimensional problem:
\begin{equation}
\label{DO}
    DO = \frac{|\bigcup_{i=1}^{|\mathcal{S}|} \bigcup_{j\ne i}^{|\mathcal{S}|}(\Omega_i\cap\Omega_j)|}{D}
\end{equation}
Then, the collaboration degree is controlled by the global optimization FEs ($GloFEs$):
\begin{equation}
\label{GloFEs}
    GloFEs = \left\{
    \begin{array}{ll}
    0 & \text{if } DO=0 \\
    (0.2 + \frac{4}{5} DO)TFEs & \text{if } DO \neq 0
    \end{array}
    \right.
\end{equation} 
 DOV algorithm \cite{meselhi2022decomposition} suggests using the mean optimization value of overlapping variables across subspaces as their optimization value to solve the coupling issue. We further calculate the weighted sum of the optimization values by computing the contribution of each subspace: 
 \begin{equation}
 \label{adjust}
     gbest_i[\gamma]=\sum_{j=i-1}^{i}\frac{\Delta_j}{\Delta_i + \Delta_{i-1}} gbest_j[\gamma] \text{ where } \gamma = \bigcap_{j=i-1}^{i} \mathcal{S}[j]
 \end{equation}
 where $\Delta_i$ is the difference in the fitness of the global best point ($gbest$) before and after the optimization of $\Omega_i$ and $gbest_i = gbest[S[i]]$. 

Pseudocode \ref{HGCC} shows the flow of HCC. After obtaining the ideal decomposition $\mathcal{S}$ from RDDSM, the values of $DO$ and $GloFEs$ can be calculated based on equation \ref{DO} and \ref{GloFEs}. Subsequently, the problem is optimized using NDA before entering the CC framework (Lines 1-5). After entering the CC framework, the remaining FEs are evenly distributed to each subspace, and then each subspace is optimized one by one (Lines 6-7). Any algorithm from the EC can be chosen as the optimizer (OPT) in HCC. Once the optimized values stagnate for 100 individuals or after the assigned $subFEs$ are exhausted, the new $gbest_i$ and other results are obtained (Lines 8-11). Then, based on equation \ref{adjust}, the $gbest_i$ and $\Omega_i$ are adjusted (Line 12). The adjustment of $\Omega_i$ simply involves replacing $gbest_i$ in equation \ref{adjust} with $\Omega_i$. The termination condition of the CC framework is when either the remaining FEs reach 0 or the target optimization value is achieved.

\section{Experiment}
\begin{table*}[t]
  \centering
  \caption{Comparing HCC-ES with comparison algorithms on AOB.}
  \vspace{-0.4cm}
  \resizebox{1\textwidth}{!}{
    \begin{tabular}{|c|cccccccccccccccccccccc|}
    \hline
    \multirow{3}[6]{*}{} & \multicolumn{12}{c||}{CC}                                          & \multicolumn{8}{c||}{NDAs}                   & \multicolumn{2}{c|}{HCC} \\
\cline{2-23}          & \multicolumn{3}{c|}{DG2-CMAES} & \multicolumn{3}{c|}{Random-CMAES} & \multicolumn{3}{c|}{RDG3-CMAES} & \multicolumn{3}{c||}{RDDSM-CMAES} & \multicolumn{2}{c|}{Sep-CMAES} & \multicolumn{2}{c|}{LM-MA-ES} & \multicolumn{2}{c|}{LMCMA} & \multicolumn{2}{c||}{MM-ES} & \multicolumn{2}{c|}{HCC-ES} \\
\cline{2-23}          & \multicolumn{1}{c|}{Perf} & \multicolumn{1}{c|}{Time} & \multicolumn{1}{c|}{Acc} & \multicolumn{1}{c|}{Perf} & \multicolumn{1}{c|}{Time} & \multicolumn{1}{c|}{Acc} & \multicolumn{1}{c|}{Perf} & \multicolumn{1}{c|}{Time} & \multicolumn{1}{c|}{Acc} & \multicolumn{1}{c|}{Perf} & \multicolumn{1}{c|}{Time} & \multicolumn{1}{c||}{Acc} & \multicolumn{1}{c|}{Perf} & \multicolumn{1}{c|}{Time} & \multicolumn{1}{c|}{Perf} & \multicolumn{1}{c|}{Time} & \multicolumn{1}{c|}{Perf} & \multicolumn{1}{c|}{Time} & \multicolumn{1}{c|}{Perf} & \multicolumn{1}{c||}{Time} & \multicolumn{1}{c|}{Perf} & Time \\
    \hline
    E1    & 3.15E6 ($\approx$)& \multirow{2}[1]{*}{9.01E2} & \multicolumn{1}{c|}{\multirow{2}[1]{*}{100\%}} & 9.80E11 (+) & \multirow{2}[1]{*}{6.94E2} & \multicolumn{1}{c|}{\multirow{2}[1]{*}{11.20\%}} & 3.13E6 ($\approx$)& \multirow{2}[1]{*}{7.95E2} & \multicolumn{1}{c|}{\multirow{2}[1]{*}{100\%}} & 2.88E6 ($\approx$)& \multirow{2}[1]{*}{8.09E2} & \multicolumn{1}{c||}{\multirow{2}[1]{*}{100\%}} & 5.62E8 (+)& \multicolumn{1}{c|}{\multirow{2}[1]{*}{6.38E2}} & \multicolumn{1}{l}{3.81E8 (+)}& \multicolumn{1}{c|}{\multirow{2}[1]{*}{1.57E3}} & 8.69E7 (+)& \multicolumn{1}{c|}{\multirow{2}[1]{*}{7.24E2}} & 1.60E7 (+)& \multicolumn{1}{c||}{\multirow{2}[1]{*}{6.57E2}} & \textbf{2.84E6} & \multirow{2}[1]{*}{7.62E2} \\
          & $\pm$2.14E5 &       & \multicolumn{1}{c|}{} & $\pm$1.53E11 &       & \multicolumn{1}{c|}{} & $\pm$2.43E5 &       & \multicolumn{1}{c|}{} & $\pm$3.12E5 &       & \multicolumn{1}{c||}{} & $\pm$4.51E4 & \multicolumn{1}{c|}{} & \multicolumn{1}{l}{$\pm$6.44E4} & \multicolumn{1}{c|}{} & \multicolumn{1}{l}{$\pm$3.43E3} & \multicolumn{1}{c|}{} & $\pm$4.21E3 & \multicolumn{1}{c||}{} & \textbf{$\pm$2.35E5} &  \\
    E2    & 2.27E11(+) & \multirow{2}[0]{*}{1.85E4} & \multicolumn{1}{c|}{\multirow{2}[0]{*}{0\%}} & 3.56E11(+) & \multirow{2}[0]{*}{7.09E2} & \multicolumn{1}{c|}{\multirow{2}[0]{*}{11.78\%}} & 6.72E8(+) & \multirow{2}[0]{*}{8.28E2} & \multicolumn{1}{c|}{\multirow{2}[0]{*}{98.63\%}} & 6.34E8(+) & \multirow{2}[0]{*}{8.34E2} & \multicolumn{1}{c||}{\multirow{2}[0]{*}{100\%}} & 5.43E8(+) & \multicolumn{1}{c|}{\multirow{2}[0]{*}{6.61E2}} & 3.40E8 (+)& \multicolumn{1}{c|}{\multirow{2}[0]{*}{1.68E3}} & 9.54E7(+) & \multicolumn{1}{c|}{\multirow{2}[0]{*}{7.30E2}} & 1.77E7(+) & \multicolumn{1}{c||}{\multirow{2}[0]{*}{6.15E2}} & \textbf{6.87E6} & \multirow{2}[0]{*}{7.72E2} \\
          & $\pm$1.31E11 &       & \multicolumn{1}{c|}{} & $\pm$1.01E11 &       & \multicolumn{1}{c|}{} & $\pm$4.15E6 &       & \multicolumn{1}{c|}{} & $\pm$5.24E6 &       & \multicolumn{1}{c||}{} & $\pm$6.23E4 & \multicolumn{1}{c|}{} & $\pm$2.39E4 & \multicolumn{1}{c|}{} & $\pm$2.77E3 & \multicolumn{1}{c|}{} & $\pm$5.78E3 & \multicolumn{1}{c||}{} & \textbf{$\pm$4.73E4} &  \\
    E3    & 2.38E11 (+) & \multirow{2}[0]{*}{1.81E4} & \multicolumn{1}{c|}{\multirow{2}[0]{*}{0\%}} & 5.45E11 (+) & \multirow{2}[0]{*}{7.34E2} & \multicolumn{1}{c|}{\multirow{2}[0]{*}{11.35\%}} & 2.02E8 (+) & \multirow{2}[0]{*}{8.38E2} & \multicolumn{1}{c|}{\multirow{2}[0]{*}{95.74\%}} & 2.07E8 (+) & \multirow{2}[0]{*}{8.50E2} & \multicolumn{1}{c||}{\multirow{2}[0]{*}{100\%}} & 6.97E8 (+)& \multicolumn{1}{c|}{\multirow{2}[0]{*}{6.69E2}} & 4.32E8 (+)& \multicolumn{1}{c|}{\multirow{2}[0]{*}{1.86E3}} & 1.40E8 (+)& \multicolumn{1}{c|}{\multirow{2}[0]{*}{8.70E2}} & 1.99E7 (+)& \multicolumn{1}{c||}{\multirow{2}[0]{*}{6.83E2}} & \textbf{1.60E7} & \multirow{2}[0]{*}{7.87E2} \\
          & $\pm$2.03E11 &       & \multicolumn{1}{c|}{} & v$\pm$2.78E11 &       & \multicolumn{1}{c|}{} & $\pm$8.46E6 &       & \multicolumn{1}{c|}{} & $\pm$2.53E6 &       & \multicolumn{1}{c||}{} & $\pm$6.84E4 & \multicolumn{1}{c|}{} & $\pm$3.42E4 & \multicolumn{1}{c|}{} & $\pm$9.37E4 & \multicolumn{1}{c|}{} & $\pm$7.55E3 & \multicolumn{1}{c||}{} & \textbf{$\pm$5.62E5} &  \\
    E4    & 2.70E11 (+)& \multirow{2}[0]{*}{2.00E4} & \multicolumn{1}{c|}{\multirow{2}[0]{*}{0\%}} & 4.75E11 (+)& \multirow{2}[0]{*}{7.38E2} & \multicolumn{1}{c|}{\multirow{2}[0]{*}{10.41\%}} & 6.35E9 (+)& \multirow{2}[0]{*}{8.36E2} & \multicolumn{1}{c|}{\multirow{2}[0]{*}{92.69\%}} & 4.86E9 (+)& \multirow{2}[0]{*}{8.34E2} & \multicolumn{1}{c||}{\multirow{2}[0]{*}{100\%}} & 4.89E8 (+)& \multicolumn{1}{c|}{\multirow{2}[0]{*}{7.14E2}} & 4.89E8 (+)& \multicolumn{1}{c|}{\multirow{2}[0]{*}{1.80E3}} & 1.13E8 (+)& \multicolumn{1}{c|}{\multirow{2}[0]{*}{7.87E2}} & \textbf{1.90E7} (-) & \multicolumn{1}{c||}{\multirow{2}[0]{*}{6.63E2}} & 2.26E7 & \multirow{2}[0]{*}{7.80E2} \\
          & $\pm$7.93E10 &       & \multicolumn{1}{c|}{} & $\pm$3.44E11 &       & \multicolumn{1}{c|}{} & $\pm$5.10E6 &       & \multicolumn{1}{c|}{} & $\pm$4.81E7 &       & \multicolumn{1}{c||}{} & $\pm$7.23E4 & \multicolumn{1}{c|}{} & $\pm$5.53E4 & \multicolumn{1}{c|}{} & $\pm$2.77E4 & \multicolumn{1}{c|}{} & \textbf{$\pm$5.45E3} & \multicolumn{1}{c||}{} & $\pm$3.04E5 &  \\
    E5    & 2.86E11 (+)& \multirow{2}[0]{*}{1.89E4} & \multicolumn{1}{c|}{\multirow{2}[0]{*}{0\%}} & 7.23E11 (+)& \multirow{2}[0]{*}{7.50E2} & \multicolumn{1}{c|}{\multirow{2}[0]{*}{10.68\%}} & 1.29E10 (+)& \multirow{2}[0]{*}{8.46E2} & \multicolumn{1}{c|}{\multirow{2}[0]{*}{90.11\%}} & 1.10E10 (+)& \multirow{2}[0]{*}{8.47E2} & \multicolumn{1}{c||}{\multirow{2}[0]{*}{100\%}} & 7.45E8 (+)& \multicolumn{1}{c|}{\multirow{2}[0]{*}{6.55E2}} & 4.61E8 (+)& \multicolumn{1}{c|}{\multirow{2}[0]{*}{2.05E3}} & 1.63E8 (+)& \multicolumn{1}{c|}{\multirow{2}[0]{*}{7.58E2}} & 2.48E7 (+) & \multicolumn{1}{c||}{\multirow{2}[0]{*}{7.13E2}} & \textbf{7.76E6} & \multirow{2}[0]{*}{7.88E2} \\
          & $\pm$1.75E11 &       & \multicolumn{1}{c|}{} & $\pm$2.53E11 &       & \multicolumn{1}{c|}{} & $\pm$4.11E7 &       & \multicolumn{1}{c|}{} & $\pm$6.23E7 &       & \multicolumn{1}{c||}{} & $\pm$5.31E4 & \multicolumn{1}{c|}{} & $\pm$2.31E4 & \multicolumn{1}{c|}{} & $\pm$3.09E4 & \multicolumn{1}{c|}{} & $\pm$4.29E3 & \multicolumn{1}{c||}{} & $\pm$\textbf{2.13E5} &  \\
    E6    & 3.20E11 (+)& \multirow{2}[0]{*}{1.94E4} & \multicolumn{1}{c|}{\multirow{2}[0]{*}{0\%}} & 5.19E11 (+)& \multirow{2}[0]{*}{7.61E2} & \multicolumn{1}{c|}{\multirow{2}[0]{*}{11.34\%}} & 3.88E10 (+)& \multirow{2}[0]{*}{1.79E3} & \multicolumn{1}{c|}{\multirow{2}[0]{*}{87.39\%}} & 3.42E10 (+)& \multirow{2}[0]{*}{8.94E2} & \multicolumn{1}{c||}{\multirow{2}[0]{*}{100\%}} & 7.92E8 (+)& \multicolumn{1}{c|}{\multirow{2}[0]{*}{7.35E2}} & 4.71E8 (+)& \multicolumn{1}{c|}{\multirow{2}[0]{*}{2.11E3}} & 1.44E8 (+)& \multicolumn{1}{c|}{\multirow{2}[0]{*}{9.35E2}} & \textbf{2.62E7} (-)& \multicolumn{1}{c||}{\multirow{2}[0]{*}{8.22E2}} & 4.32E7 & \multirow{2}[0]{*}{7.98E2} \\
          & $\pm$9.22E10 &       & \multicolumn{1}{c|}{} & $\pm$1.21E11 &       & \multicolumn{1}{c|}{} & $\pm$3.22E7 &       & \multicolumn{1}{c|}{} & $\pm$1.69E8 &       & \multicolumn{1}{c||}{} & $\pm$3.07E4 & \multicolumn{1}{c|}{} & $\pm$4.02E4 & \multicolumn{1}{c|}{} & $\pm$4.14E4 & \multicolumn{1}{c|}{} & \textbf{$\pm$2.37E3} & \multicolumn{1}{c||}{} & $\pm$1.94E5 &  \\
    \hline
    S1    & 1.98E-3 ($\approx$)& \multirow{2}[0]{*}{1.12E3} & \multicolumn{1}{c|}{\multirow{2}[0]{*}{100\%}} & 9.67E8 (+)& \multirow{2}[0]{*}{8.82E2} & \multicolumn{1}{c|}{\multirow{2}[0]{*}{11.30\%}} & 1.37E-3 ($\approx$)& \multirow{2}[0]{*}{9.51E2} & \multicolumn{1}{c|}{\multirow{2}[0]{*}{100\%}} & 1.46E-3 ($\approx$)& \multirow{2}[0]{*}{1.00E3} & \multicolumn{1}{c||}{\multirow{2}[0]{*}{100\%}} & 5.06E6 (+)& \multicolumn{1}{c|}{\multirow{2}[0]{*}{8.62E2}} & 2.22E5 (+)& \multicolumn{1}{c|}{\multirow{2}[0]{*}{2.06E3}} & 5.50E5 (+)& \multicolumn{1}{c|}{\multirow{2}[0]{*}{9.48E2}} & 1.67E4 (+)& \multicolumn{1}{c||}{\multirow{2}[0]{*}{8.95E2}} & \textbf{1.92E-3} & \multirow{2}[0]{*}{9.46E2} \\
          & $\pm$5.18E-4 &       & \multicolumn{1}{c|}{} & $\pm$1.13E8 &       & \multicolumn{1}{c|}{} & $\pm$4.23E-4 &       & \multicolumn{1}{c|}{} & $\pm$5.29E-4 &       & \multicolumn{1}{c||}{} & $\pm$1.66E2 & \multicolumn{1}{c|}{} & $\pm$3.51E1 & \multicolumn{1}{c|}{} & $\pm$3.05E1 & \multicolumn{1}{c|}{} & $\pm$2.14E1 & \multicolumn{1}{c||}{} & \textbf{$\pm$3.82E-4} &  \\
    S2    & 4.67E8 (+)& \multirow{2}[0]{*}{1.83E4} & \multicolumn{1}{c|}{\multirow{2}[0]{*}{0\%}} & 1.04E9 (+)& \multirow{2}[0]{*}{8.96E2} & \multicolumn{1}{c|}{\multirow{2}[0]{*}{10.79\%}} & 7.92E5 (+)& \multirow{2}[0]{*}{9.83E2} & \multicolumn{1}{c|}{\multirow{2}[0]{*}{98.63\%}} & 8.10E4 (+)& \multirow{2}[0]{*}{1.01E3} & \multicolumn{1}{c||}{\multirow{2}[0]{*}{100\%}} & 2.63E6 (+)& \multicolumn{1}{c|}{\multirow{2}[0]{*}{8.35E2}} & 2.20E5 (+)& \multicolumn{1}{c|}{\multirow{2}[0]{*}{1.91E3}} & 7.76E5 (+)& \multicolumn{1}{c|}{\multirow{2}[0]{*}{9.89E2}} & 1.30E4 (+)& \multicolumn{1}{c||}{\multirow{2}[0]{*}{8.58E2}} & \textbf{5.58E3} & \multirow{2}[0]{*}{9.62E2} \\
          & $\pm$5.28E7 &       & \multicolumn{1}{c|}{} & $\pm$$\pm$3.99E8 &       & \multicolumn{1}{c|}{} & $\pm$5.26E4 &       & \multicolumn{1}{c|}{} & $\pm$5.43E3 &       & \multicolumn{1}{c||}{} & $\pm$2.15E2 & \multicolumn{1}{c|}{} & $\pm$1.41E1 & \multicolumn{1}{c|}{} & $\pm$2.23E1 & \multicolumn{1}{c|}{} & $\pm$7.54E1 & \multicolumn{1}{c||}{} & \textbf{$\pm$3.21E1}&  \\
    S3    & 4.57E8 (+)& \multirow{2}[0]{*}{1.69E4} & \multicolumn{1}{c|}{\multirow{2}[0]{*}{0\%}} & 9.97E8 (+)& \multirow{2}[0]{*}{9.15E2} & \multicolumn{1}{c|}{\multirow{2}[0]{*}{10.79\%}} & 7.60E6 (+)& \multirow{2}[0]{*}{9.99E2} & \multicolumn{1}{c|}{\multirow{2}[0]{*}{95.63\%}} & 2.48E6 (+)& \multirow{2}[0]{*}{1.02E3} & \multicolumn{1}{c||}{\multirow{2}[0]{*}{100\%}} & 4.14E6 (+)& \multicolumn{1}{c|}{\multirow{2}[0]{*}{8.85E2}} & 2.43E5 (+)& \multicolumn{1}{c|}{\multirow{2}[0]{*}{2.01E3}} & 4.89E5 (+)& \multicolumn{1}{c|}{\multirow{2}[0]{*}{1.06E3}} & 1.91E4 (+)& \multicolumn{1}{c||}{\multirow{2}[0]{*}{9.36E2}} & \textbf{9.72E3} & \multirow{2}[0]{*}{9.69E2} \\
          & $\pm$3.08E7 &       & \multicolumn{1}{c|}{} & $\pm$1.21E8 &       & \multicolumn{1}{c|}{} & $\pm$6.13E5 &       & \multicolumn{1}{c|}{} & $\pm$4.60E5 &       & \multicolumn{1}{c||}{} & $\pm$3.43E2 & \multicolumn{1}{c|}{} & $\pm$3.12E1 & \multicolumn{1}{c|}{} & $\pm$5.13E1 & \multicolumn{1}{c|}{} & $\pm$9.47E1 & \multicolumn{1}{c||}{} & \textbf{$\pm$4.75E1} &  \\
    S4    & 5.43E8 (+)& \multirow{2}[0]{*}{1.58E4} & \multicolumn{1}{c|}{\multirow{2}[0]{*}{0\%}} & 8.44E8 (+)& \multirow{2}[0]{*}{9.19E2} & \multicolumn{1}{c|}{\multirow{2}[0]{*}{10.50\%}} & 2.03E8 (+)& \multirow{2}[0]{*}{1.00E3} & \multicolumn{1}{c|}{\multirow{2}[0]{*}{93.15\%}} & 2.88E7 (+)& \multirow{2}[0]{*}{1.00E3} & \multicolumn{1}{c||}{\multirow{2}[0]{*}{100\%}} & 3.14E6 (+)& \multicolumn{1}{c|}{\multirow{2}[0]{*}{1.01E3}} & 2.54E5 (+)& \multicolumn{1}{c|}{\multirow{2}[0]{*}{2.06E3}} & 5.35E5 (+)& \multicolumn{1}{c|}{\multirow{2}[0]{*}{1.05E3}} & 1.50E4 (+)& \multicolumn{1}{c||}{\multirow{2}[0]{*}{1.02E3}} & \textbf{1.24E3} & \multirow{2}[0]{*}{9.50E2} \\
          & $\pm$2.84E7 &       & \multicolumn{1}{c|}{} & $\pm$2.37E8 &       & \multicolumn{1}{c|}{} & $\pm$4.27E7 &       & \multicolumn{1}{c|}{} & $\pm$9.25E6 &       & \multicolumn{1}{c||}{} & $\pm$5.67E2 & \multicolumn{1}{c|}{} & $\pm$4.27E1 & \multicolumn{1}{c|}{} & $\pm$7.42E1 & \multicolumn{1}{c|}{} & $\pm$1.05E2 & \multicolumn{1}{c||}{} & \textbf{$\pm$5.31E1} &  \\
    S5    & 6.83E8 (+)& \multirow{2}[0]{*}{1.60E4} & \multicolumn{1}{c|}{\multirow{2}[0]{*}{0\%}} & 1.32E9 (+)& \multirow{2}[0]{*}{9.38E2} & \multicolumn{1}{c|}{\multirow{2}[0]{*}{10.77\%}} & 8.84E7 (+)& \multirow{2}[0]{*}{1.01E3} & \multicolumn{1}{c|}{\multirow{2}[0]{*}{90.56\%}} & 6.77E6 (+)& \multirow{2}[0]{*}{1.01E3} & \multicolumn{1}{c||}{\multirow{2}[0]{*}{100\%}} & 2.14E6 (+)& \multicolumn{1}{c|}{\multirow{2}[0]{*}{9.55E2}} & 3.01E5 (+)& \multicolumn{1}{c|}{\multirow{2}[0]{*}{2.18E3}} & 1.32E6 (+)& \multicolumn{1}{c|}{\multirow{2}[0]{*}{1.12E3}} & 1.87E4 (+)& \multicolumn{1}{c||}{\multirow{2}[0]{*}{9.95E2}} & \textbf{9.23E3} & \multirow{2}[0]{*}{9.46E2} \\
          & $\pm$7.42E7 &       & \multicolumn{1}{c|}{} & $\pm$5.91E8 &       & \multicolumn{1}{c|}{} & $\pm$3.54E6 &       & \multicolumn{1}{c|}{} & $\pm$4.60E5 &       & \multicolumn{1}{c||}{} & $\pm$3.48E2 & \multicolumn{1}{c|}{} & $\pm$3.65E1 & \multicolumn{1}{c|}{} & $\pm$9.13E1 & \multicolumn{1}{c|}{} & $\pm$8.53E1 & \multicolumn{1}{c||}{} & \textbf{$\pm$3.95E2} &  \\
    S6    & 7.01E8 (+)& \multirow{2}[0]{*}{2.06E4} & \multicolumn{1}{c|}{\multirow{2}[0]{*}{0\%}} & 1.06E9 (+)& \multirow{2}[0]{*}{9.56E2} & \multicolumn{1}{c|}{\multirow{2}[0]{*}{11.09\%}} & 9.01E7 (+)& \multirow{2}[0]{*}{1.10E3} & \multicolumn{1}{c|}{\multirow{2}[0]{*}{86.97\%}} & 6.01E6 (+)& \multirow{2}[0]{*}{1.12E3} & \multicolumn{1}{c||}{\multirow{2}[0]{*}{100\%}} & 2.86E6 (+)& \multicolumn{1}{c|}{\multirow{2}[0]{*}{1.05E3}} & 2.26E5 (+)& \multicolumn{1}{c|}{\multirow{2}[0]{*}{2.61E3}} & 7.38E5 (+)& \multicolumn{1}{c|}{\multirow{2}[0]{*}{1.11E3}} & \textbf{1.33E4} (-)& \multicolumn{1}{c||}{\multirow{2}[0]{*}{9.93E2}} & 6.65E4 (+)& \multirow{2}[0]{*}{9.61E2} \\
          & $\pm$6.12E7 &       & \multicolumn{1}{c|}{} & $\pm$3.13E8 &       & \multicolumn{1}{c|}{} & $\pm$5.27E6 &       & \multicolumn{1}{c|}{} & $\pm$5.73E5 &       & \multicolumn{1}{c||}{} & $\pm$5.73E2 & \multicolumn{1}{c|}{} & $\pm$4.42E1 & \multicolumn{1}{c|}{} & $\pm$7.82E1 & \multicolumn{1}{c|}{} & \textbf{$\pm$7.19E1} & \multicolumn{1}{c||}{} & $\pm$7.36E2 &  \\
    \hline
    R1    & 1.87E5 ($\approx$)& \multirow{2}[0]{*}{1.79E3} & \multicolumn{1}{c|}{\multirow{2}[0]{*}{100\%}} & 4.07E8 (+)& \multirow{2}[0]{*}{1.01E3} & \multicolumn{1}{c|}{\multirow{2}[0]{*}{10.90\%}} & 1.74E5 ($\approx$)& \multirow{2}[0]{*}{1.83E3} & \multicolumn{1}{c|}{\multirow{2}[0]{*}{100\%}} & 1.76E5 ($\approx$)& \multirow{2}[0]{*}{1.81E3} & \multicolumn{1}{c||}{\multirow{2}[0]{*}{100\%}} & 2.03E5 (+)& \multicolumn{1}{c|}{\multirow{2}[0]{*}{9.88E2}} & 7.92E5 (+)& \multicolumn{1}{c|}{\multirow{2}[0]{*}{2.68E3}} & 3.55E6 (+)& \multicolumn{1}{c|}{\multirow{2}[0]{*}{1.13E3}} & 1.36E6 (+)& \multicolumn{1}{c||}{\multirow{2}[0]{*}{1.06E3}} & \textbf{1.74E5} & \multirow{2}[0]{*}{1.23E3} \\
          & $\pm$3.11E3 &       & \multicolumn{1}{c|}{} & $\pm$1.13E8 &       & \multicolumn{1}{c|}{} & $\pm$6.67E3 &       & \multicolumn{1}{c|}{} & $\pm$2.03E3 &       & \multicolumn{1}{c||}{} & $\pm$2.27E1 & \multicolumn{1}{c|}{} & $\pm$9.23E1 & \multicolumn{1}{c|}{} & $\pm$3.63E2 & \multicolumn{1}{c|}{} & $\pm$4.01E2 & \multicolumn{1}{c||}{} & \textbf{$\pm$8.13E3} &  \\
    R2    & 4.39E7 (+)& \multirow{2}[0]{*}{2.19E4} & \multicolumn{1}{c|}{\multirow{2}[0]{*}{0\%}} & 7.96E8 (+)& \multirow{2}[0]{*}{1.53E3} & \multicolumn{1}{c|}{\multirow{2}[0]{*}{10.89\%}} & 1.47E6 (+)& \multirow{2}[0]{*}{1.61E3} & \multicolumn{1}{c|}{\multirow{2}[0]{*}{98.63\%}} & 3.02E5 (-)& \multirow{2}[0]{*}{1.68E3} & \multicolumn{1}{c||}{\multirow{2}[0]{*}{100\%}} & \textbf{2.48E5} (-)& \multicolumn{1}{c|}{\multirow{2}[0]{*}{1.02E3}} & 8.30E5 (+)& \multicolumn{1}{c|}{\multirow{2}[0]{*}{2.18E3}} & 2.17E6 (+)& \multicolumn{1}{c|}{\multirow{2}[0]{*}{1.16E3}} & 1.39E6 (+)& \multicolumn{1}{c||}{\multirow{2}[0]{*}{1.04E3}} & 3.72E5 & \multirow{2}[0]{*}{1.25E3} \\
          & $\pm$2.04E6 &       & \multicolumn{1}{c|}{} & $\pm$2.25E8 &       & \multicolumn{1}{c|}{} & $\pm$2.83E5 &       & \multicolumn{1}{c|}{} & $\pm$1.37E4 &       & \multicolumn{1}{c||}{} & \textbf{$\pm$3.82E1} & \multicolumn{1}{c|}{} & $\pm$8.39E1 & \multicolumn{1}{c|}{} & $\pm$2.31E2 & \multicolumn{1}{c|}{} & $\pm$3.71E2 & \multicolumn{1}{c||}{} & $\pm$5.27E3 &  \\
    R3    & 4.43E7 (+)& \multirow{2}[0]{*}{1.88E4} & \multicolumn{1}{c|}{\multirow{2}[0]{*}{0\%}} & 1.06E9 (+)& \multirow{2}[0]{*}{1.73E3} & \multicolumn{1}{c|}{\multirow{2}[0]{*}{10.88\%}} & 9.57E5 (+)& \multirow{2}[0]{*}{1.49E3} & \multicolumn{1}{c|}{\multirow{2}[0]{*}{95.74\%}} & 6.16E5 (+)& \multirow{2}[0]{*}{1.52E3} & \multicolumn{1}{c||}{\multirow{2}[0]{*}{100\%}} & \textbf{3.28E5} (-)& \multicolumn{1}{c|}{\multirow{2}[0]{*}{1.11E3}} & 8.85E5 (+)& \multicolumn{1}{c|}{\multirow{2}[0]{*}{1.81E3}} & 2.97E6 (+)& \multicolumn{1}{c|}{\multirow{2}[0]{*}{1.11E3}} & 1.34E6 (+)& \multicolumn{1}{c||}{\multirow{2}[0]{*}{1.14E3}} & 5.06E5 & \multirow{2}[0]{*}{1.20E3} \\
          & $\pm$6.83E6 &       & \multicolumn{1}{c|}{} & $\pm$4.77E8 &       & \multicolumn{1}{c|}{} & $\pm$9.89E4 &       & \multicolumn{1}{c|}{} & $\pm$2.64E4 &       & \multicolumn{1}{c||}{} & \textbf{$\pm$4.13E1} & \multicolumn{1}{c|}{} & $\pm$6.55E1 & \multicolumn{1}{c|}{} & $\pm$2.52E2 & \multicolumn{1}{c|}{} & $\pm$6.57E2 & \multicolumn{1}{c||}{} & $\pm$7.26E3 &  \\
    R4    & 4.61E7 (+)& \multirow{2}[0]{*}{1.69E4} & \multicolumn{1}{c|}{\multirow{2}[0]{*}{0\%}} & 1.39E9 (+)& \multirow{2}[0]{*}{1.43E3} & \multicolumn{1}{c|}{\multirow{2}[0]{*}{11.14\%}} & 2.60E6 (+)& \multirow{2}[0]{*}{1.41E3} & \multicolumn{1}{c|}{\multirow{2}[0]{*}{92.51\%}} & 7.93E5 (+)& \multirow{2}[0]{*}{1.48E3} & \multicolumn{1}{c||}{\multirow{2}[0]{*}{100\%}} & \textbf{3.20E5} (-)& \multicolumn{1}{c|}{\multirow{2}[0]{*}{1.10E3}} & 8.97E5 (+)& \multicolumn{1}{c|}{\multirow{2}[0]{*}{2.52E3}} & 2.36E6 (+)& \multicolumn{1}{c|}{\multirow{2}[0]{*}{1.29E3}} & 1.48E6 (+)& \multicolumn{1}{c||}{\multirow{2}[0]{*}{1.12E3}} & 5.59E5 & \multirow{2}[0]{*}{1.17E3} \\
          & $\pm$2.33E6 &       & \multicolumn{1}{c|}{} & $\pm$2.25E8 &       & \multicolumn{1}{c|}{} & $\pm$2.83E5 &       & \multicolumn{1}{c|}{} & $\pm$1.37E4 &       & \multicolumn{1}{c||}{} & \textbf{$\pm$3.37E1} & \multicolumn{1}{c|}{} & $\pm$4.23E1 & \multicolumn{1}{c|}{} & $\pm$3.53E2 & \multicolumn{1}{c|}{} & $\pm$2.68E2 & \multicolumn{1}{c||}{} & 6.77E3 &  \\
    R5    & 5.25E7 (+)& \multirow{2}[0]{*}{2.52E4 } & \multicolumn{1}{c|}{\multirow{2}[0]{*}{0\%}} & 6.84E8 (+)& \multirow{2}[0]{*}{1.41E3} & \multicolumn{1}{c|}{\multirow{2}[0]{*}{10.59\%}} & 2.05E6 (+)& \multirow{2}[0]{*}{1.49E3} & \multicolumn{1}{c|}{\multirow{2}[0]{*}{90.03\%}} & 1.25E6 (+)& \multirow{2}[0]{*}{1.51E3} & \multicolumn{1}{c||}{\multirow{2}[0]{*}{100\%}} & \textbf{2.48E5} (-)& \multicolumn{1}{c|}{\multirow{2}[0]{*}{1.09E3}} & 9.50E5 (+)& \multicolumn{1}{c|}{\multirow{2}[0]{*}{3.18E3}} & 4.05E6 (+)& \multicolumn{1}{c|}{\multirow{2}[0]{*}{1.27E3}} & 1.52E6 (+)& \multicolumn{1}{c||}{\multirow{2}[0]{*}{1.06E3}} & 6.84E5 & \multirow{2}[0]{*}{1.20E3} \\
          & $\pm$4.11E6 &       & \multicolumn{1}{c|}{} & $\pm$1.56E8 &       & \multicolumn{1}{c|}{} & $\pm$3.21E5 &       & \multicolumn{1}{c|}{} & $\pm$8.98E4 &       & \multicolumn{1}{c||}{} & \textbf{$\pm$1.92E1} & \multicolumn{1}{c|}{} & $\pm$2.19E1 & \multicolumn{1}{c|}{} & $\pm$2.84E2 & \multicolumn{1}{c|}{} & $\pm$3.55E2 & \multicolumn{1}{c||}{} & $\pm$5.76E3 &  \\
    R6    & 5.86E7 (+)& \multirow{2}[0]{*}{1.69E4} & \multicolumn{1}{c|}{\multirow{2}[0]{*}{0\%}} & 1.41E9 (+)& \multirow{2}[0]{*}{1.53E3} & \multicolumn{1}{c|}{\multirow{2}[0]{*}{10.84\%}} & 3.85E6 (+)& \multirow{2}[0]{*}{1.54E3} & \multicolumn{1}{c|}{\multirow{2}[0]{*}{87.23\%}} & 1.30E6 (+)& \multirow{2}[0]{*}{1.64E3} & \multicolumn{1}{c||}{\multirow{2}[0]{*}{100\%}} & \textbf{2.66E5} (-)& \multicolumn{1}{c|}{\multirow{2}[0]{*}{1.13E3}} & 9.57E5 (+)& \multicolumn{1}{c|}{\multirow{2}[0]{*}{3.30E3}} & 3.03E6 (+)& \multicolumn{1}{c|}{\multirow{2}[0]{*}{1.26E3}} & 1.68E6 (+)& \multicolumn{1}{c||}{\multirow{2}[0]{*}{1.12E3}} & 8.15E5 & \multirow{2}[0]{*}{1.23E3} \\
          & $\pm$3.50E6 &       & \multicolumn{1}{c|}{} & $\pm$4.26E8 &       & \multicolumn{1}{c|}{} & $\pm$2.57E5 &       & \multicolumn{1}{c|}{} & $\pm$3.47E5 &       & \multicolumn{1}{c||}{} & \textbf{$\pm$4.35E1} & \multicolumn{1}{c|}{} & $\pm$3.53E1 & \multicolumn{1}{c|}{} & $\pm$3.37E2 & \multicolumn{1}{c|}{} & $\pm$1.16E2 & \multicolumn{1}{c||}{} & $\pm$2.30E4 &  \\
    \hline
    A1    & 7.76E4 ($\approx$)& \multirow{2}[0]{*}{1.85E3} & \multicolumn{1}{c|}{\multirow{2}[0]{*}{100\%}} & 7.93E4 (+)& \multirow{2}[0]{*}{1.91E3} & \multicolumn{1}{c|}{\multirow{2}[0]{*}{10.70\%}} & 7.76E4 ($\approx$)& \multirow{2}[0]{*}{1.84E3} & \multicolumn{1}{c|}{\multirow{2}[0]{*}{100\%}} & 7.76E4 ($\approx$)& \multirow{2}[0]{*}{1.42E3} & \multicolumn{1}{c||}{\multirow{2}[0]{*}{100\%}} & 7.81E4 (+)& \multicolumn{1}{c|}{\multirow{2}[0]{*}{1.11E3}} & 7.84E4 (+)& \multicolumn{1}{c|}{\multirow{2}[0]{*}{3.06E3}} & 7.94E4 (+)& \multicolumn{1}{c|}{\multirow{2}[0]{*}{1.23E3}} & 7.94E4 (+)& \multicolumn{1}{c||}{\multirow{2}[0]{*}{1.13E3}} & \textbf{7.76E4} & \multirow{2}[0]{*}{1.20E3} \\
          & $\pm$1.03E2 &       & \multicolumn{1}{c|}{} & $\pm$1.13E2 &       & \multicolumn{1}{c|}{} & $\pm$9.04E2 &       & \multicolumn{1}{c|}{} & $\pm$1.07E1 &       & \multicolumn{1}{c||}{} & $\pm$1.76E0 & \multicolumn{1}{c|}{} & $\pm$7.31E0 & \multicolumn{1}{c|}{} & $\pm$1.79E2 & \multicolumn{1}{c|}{} & $\pm$2.78E2 & \multicolumn{1}{c||}{} & \textbf{$\pm$1.25E1} &  \\
    A2    & 8.09E4 (+)& \multirow{2}[0]{*}{1.62E4} & \multicolumn{1}{c|}{\multirow{2}[0]{*}{0\%}} & 7.97E4 (+)& \multirow{2}[0]{*}{1.90E3} & \multicolumn{1}{c|}{\multirow{2}[0]{*}{10.79\%}} & 8.07E4 (+)& \multirow{2}[0]{*}{2.01E3} & \multicolumn{1}{c|}{\multirow{2}[0]{*}{98.63\%}} & 7.82E4 ($\approx$)& \multirow{2}[0]{*}{1.54E3} & \multicolumn{1}{c||}{\multirow{2}[0]{*}{100\%}} & 7.81E4 ($\approx$)& \multicolumn{1}{c|}{\multirow{2}[0]{*}{1.16E3}} & 7.86E4 (+)& \multicolumn{1}{c|}{\multirow{2}[0]{*}{3.08E3}} & 7.86E4 (+)& \multicolumn{1}{c|}{\multirow{2}[0]{*}{1.25E3}} & 7.83E4 (+)& \multicolumn{1}{c||}{\multirow{2}[0]{*}{1.12E3}} & \textbf{7.81E4} & \multirow{2}[0]{*}{1.20E3} \\
          & $\pm$7.90E2 &       & \multicolumn{1}{c|}{} & $\pm$1.25E2 &       & \multicolumn{1}{c|}{} & $\pm$8.93E2 &       & \multicolumn{1}{c|}{} & $\pm$2.33E1 &       & \multicolumn{1}{c||}{} & $\pm$2.44E0 & \multicolumn{1}{c|}{} & $\pm$1.34E2 & \multicolumn{1}{c|}{} & $\pm$2.55E1 & \multicolumn{1}{c|}{} & $\pm$9.32E0 & \multicolumn{1}{c||}{} & \textbf{$\pm$9.78E0} &  \\
    A3    & 8.10E4 (+)& \multirow{2}[0]{*}{1.68E4} & \multicolumn{1}{c|}{\multirow{2}[0]{*}{0\%}} & 7.98E4 (+)& \multirow{2}[0]{*}{1.90E3} & \multicolumn{1}{c|}{\multirow{2}[0]{*}{10.79\%}} & 8.12E4 (+)& \multirow{2}[0]{*}{2.02E3} & \multicolumn{1}{c|}{\multirow{2}[0]{*}{95.74\%}} & 7.81E4 (-)& \multirow{2}[0]{*}{1.87E3} & \multicolumn{1}{c||}{\multirow{2}[0]{*}{100\%}} & \textbf{7.80E4} (-)& \multicolumn{1}{c|}{\multirow{2}[0]{*}{1.09E3}} & 7.84E4 (-)& \multicolumn{1}{c|}{\multirow{2}[0]{*}{3.03E3}} & 7.92E4 (+)& \multicolumn{1}{c|}{\multirow{2}[0]{*}{1.27E3}} & 7.85E4 ($\approx$)& \multicolumn{1}{c||}{\multirow{2}[0]{*}{1.20E3}} & 7.86E4 & \multirow{2}[0]{*}{1.25E3} \\
          & $\pm$8.23E2 &       & \multicolumn{1}{c|}{} & $\pm$1.88E2 &       & \multicolumn{1}{c|}{} & $\pm$9.83E2 &       & \multicolumn{1}{c|}{} & $\pm$3.01E1 &       & \multicolumn{1}{c||}{} & \textbf{$\pm$3.89E0} & \multicolumn{1}{c|}{} & $\pm$8.78E1 & \multicolumn{1}{c|}{} & $\pm$1.03E2 & \multicolumn{1}{c|}{} & $\pm$1.53E1 & \multicolumn{1}{c||}{} & $\pm$2.09E1 &  \\
    A4    & 8.12E4 (+)& \multirow{2}[0]{*}{1.63E4} & \multicolumn{1}{c|}{\multirow{2}[0]{*}{0\%}} & 7.96E4 (+)& \multirow{2}[0]{*}{1.82E3} & \multicolumn{1}{c|}{\multirow{2}[0]{*}{11.14\%}} & 7.90E4 (+)& \multirow{2}[0]{*}{1.75E3} & \multicolumn{1}{c|}{\multirow{2}[0]{*}{92.69\%}} & 7.83E4 ($\approx$)& \multirow{2}[0]{*}{2.01E3} & \multicolumn{1}{c||}{\multirow{2}[0]{*}{100\%}} & 7.84E4 ($\approx$)& \multicolumn{1}{c|}{\multirow{2}[0]{*}{1.20E3}} & 7.85E4 (+)& \multicolumn{1}{c|}{\multirow{2}[0]{*}{3.10E3}} & 7.87E4 (+)& \multicolumn{1}{c|}{\multirow{2}[0]{*}{1.44E3}} & 7.89E4 (+)& \multicolumn{1}{c||}{\multirow{2}[0]{*}{1.24E3}} & \textbf{7.83E4} & \multirow{2}[0]{*}{1.24E3} \\
          & $\pm$1.22E3 &       & \multicolumn{1}{c|}{} & $\pm$1.79E2 &       & \multicolumn{1}{c|}{} & $\pm$3.13E0 &       & \multicolumn{1}{c|}{} & $\pm$1.04E1 &       & \multicolumn{1}{c||}{} & $\pm$4.27E1 & \multicolumn{1}{c|}{} & $\pm$6.90E1 & \multicolumn{1}{c|}{} & $\pm$2.63E1 & \multicolumn{1}{c|}{} & $\pm$2.54E2 & \multicolumn{1}{c||}{} & \textbf{$\pm$1.14E1} &  \\
    A5    & 8.14E4 (+)& \multirow{2}[0]{*}{1.84E4} & \multicolumn{1}{c|}{\multirow{2}[0]{*}{0\%}} & 8.00E4 (+)& \multirow{2}[0]{*}{1.84E3} & \multicolumn{1}{c|}{\multirow{2}[0]{*}{10.68\%}} & 7.99E4 (+)& \multirow{2}[0]{*}{1.35E3} & \multicolumn{1}{c|}{\multirow{2}[0]{*}{90.11\%}} & 7.85E4 ($\approx$)& \multirow{2}[0]{*}{2.02E3} & \multicolumn{1}{c||}{\multirow{2}[0]{*}{100\%}} & \textbf{7.82E4} (-)& \multicolumn{1}{c|}{\multirow{2}[0]{*}{1.23E3}} & 7.84E4 ($\approx$)& \multicolumn{1}{c|}{\multirow{2}[0]{*}{3.22E3}} & 8.06E4 (+)& \multicolumn{1}{c|}{\multirow{2}[0]{*}{1.61E3}} & 7.87E4 (+)& \multicolumn{1}{c||}{\multirow{2}[0]{*}{1.44E3}} & 7.85E4 & \multirow{2}[0]{*}{1.24E3} \\
          & $\pm$1.39E3 &       & \multicolumn{1}{c|}{} & $\pm$1.93E2 &       & \multicolumn{1}{c|}{} & $\pm$3.04E1 &       & \multicolumn{1}{c|}{} & $\pm$3.15E1 &       & \multicolumn{1}{c||}{} & \textbf{$\pm$3.42E0} & \multicolumn{1}{c|}{} & $\pm$5.43E1 & \multicolumn{1}{c|}{} & $\pm$4.59E2 & \multicolumn{1}{c|}{} & $\pm$3.17E1 & \multicolumn{1}{c||}{} & $\pm$2.35E1 &  \\
    A6    & 8.13E4 (+)& \multirow{2}[1]{*}{2.29E4} & \multicolumn{1}{c|}{\multirow{2}[1]{*}{0\%}} & 7.96E4 (+)& \multirow{2}[1]{*}{1.90E3} & \multicolumn{1}{c|}{\multirow{2}[1]{*}{10.17\%}} & 7.91E4 (+)& \multirow{2}[1]{*}{1.57E3} & \multicolumn{1}{c|}{\multirow{2}[1]{*}{87.39\%}} & 7.82E4 (+)& \multirow{2}[1]{*}{1.70E3} & \multicolumn{1}{c||}{\multirow{2}[1]{*}{100\%}} & 7.80E4 ($\approx$)& \multicolumn{1}{c|}{\multirow{2}[1]{*}{1.53E3}} & 7.84E4 (+)& \multicolumn{1}{c|}{\multirow{2}[1]{*}{3.35E3}} & 7.99E4 (+)& \multicolumn{1}{c|}{\multirow{2}[1]{*}{1.63E3}} & 7.89E4 (+)& \multicolumn{1}{c||}{\multirow{2}[1]{*}{1.49E3}} & \textbf{7.80E4} & \multirow{2}[1]{*}{1.29E3} \\
          & $\pm$1.12E3 &       & \multicolumn{1}{c|}{} & $\pm$1.57E2 &       & \multicolumn{1}{c|}{} & $\pm$2.11E0 &       & \multicolumn{1}{c|}{} & $\pm$5.23E0 &       & \multicolumn{1}{c||}{} & $\pm$7.25E0 & \multicolumn{1}{c|}{} & $\pm$3.28E1 & \multicolumn{1}{c|}{} & $\pm$6.97E2 & \multicolumn{1}{c|}{} & $\pm$4.79E1 & \multicolumn{1}{c||}{} & \textbf{$\pm$7.06E0} &  \\
    \hline
    ALL   & \multicolumn{3}{c}{20/4/0} & \multicolumn{3}{c}{24/0/0} & \multicolumn{3}{c}{20/4/0} & \multicolumn{3}{c}{15/7/2} & \multicolumn{2}{c}{14/3/8} & \multicolumn{2}{c}{22/1/1} & \multicolumn{2}{c}{24/0/0} & \multicolumn{2}{c}{20/1/3} & \multicolumn{2}{c|}{NA} \\
    \hline
    \end{tabular}%
    }
  \label{table_AOB}%
\end{table*}%

\label{section:exp}
\subsection{Auto Overlapping Benchmark}
\label{subsection:AOB}
\begin{figure}[t]
    \centering
    \includegraphics[width=0.8\linewidth]{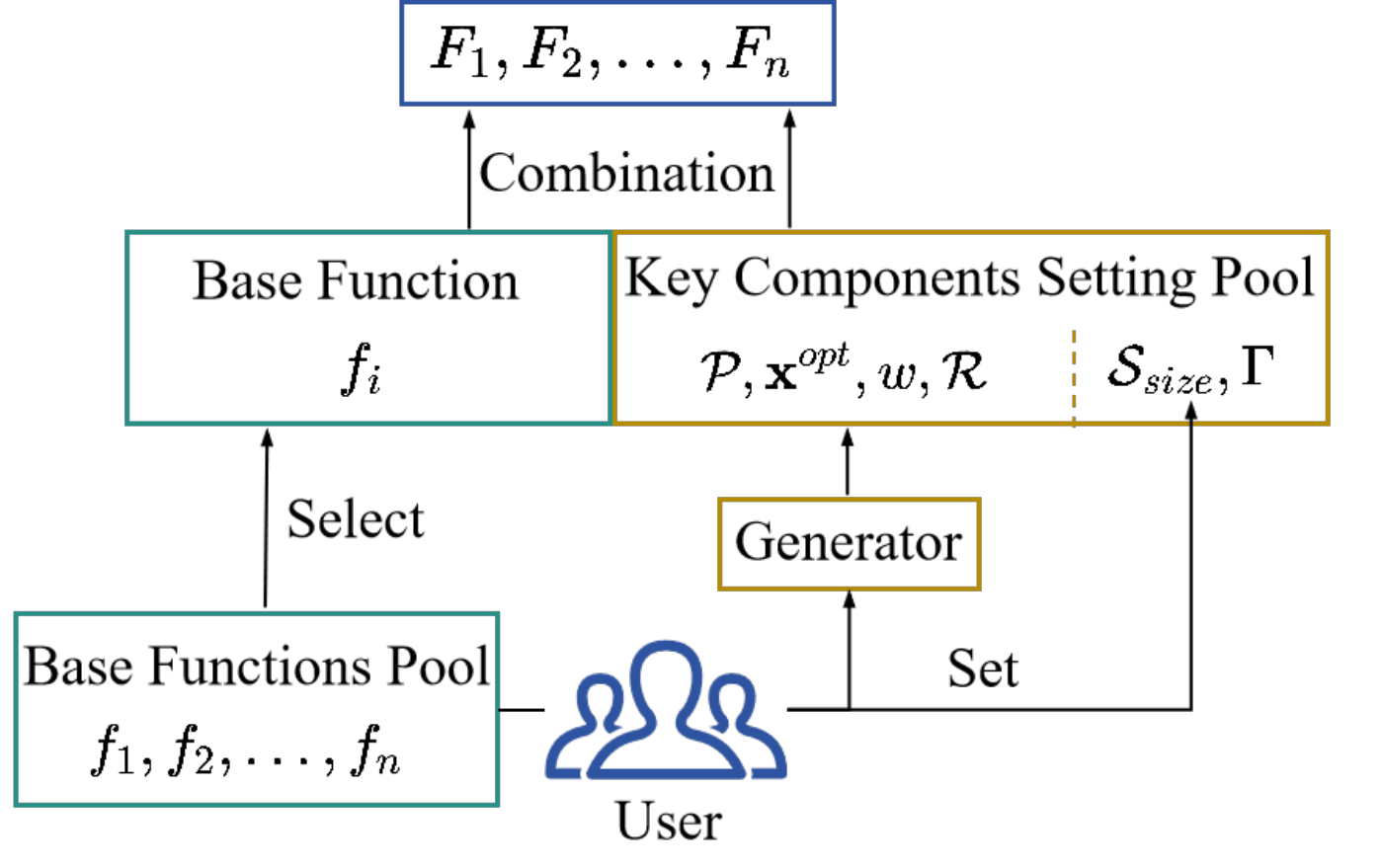}
    \caption{The key ideas of AOB.}
    \label{AOB}
    \vspace{-0.5cm}
\end{figure}

As mentioned in Section \ref{section:Background}, CEC2013LSGO consists of 15 problems, of which only 2 are overlapping. A series of novel benchmarks have been proposed to expand CEC2013LSGO \cite{xu2023large,tian2024composite,blanchard2021investigating,chen2022decomposition}. 
However, the problems in these benchmarks are usually predefined, with settings coupled together for specific problems. To conduct a more detailed analysis of the impact of overlap and facilitate the generation of overlapping problems, we propose Auto Overlapping Benchmark (AOB). AOB decoupled the key components and base functions, forming two separate pools. The problems in AOB are created by selecting elements from both pools and combining them, which allows users to freely control each component and analyze its impact. Figure \ref{AOB} illustrates the key ideas of AOB. Besides, AOB is also user-friendly, providing an automated process to generate custom functions as needed. The setting of $\mathcal{P}$, $\mathbf{x}^{opt}$, $w$, and $\mathcal{R}$ are determined by a generator, ensuring they can be set uniformly. Therefore, the users only need to select the $f$ from base functions pool and set generator, $\mathcal{S}_{size}$ and $\Gamma$, the problem will be generated automatically. 

Table \ref{table:AOB} presents the specific details of AOB.
In this work, we generate overlapping problems based on the base functions of CEC2013LSGO—Schwefel, Elliptic, Rastrigin, and Ackley—to form AOB. Since we focus on the impact of overlap on optimization performance, we generate $\Gamma_1$, $\Gamma_2$, $\Gamma_3$, $\Gamma_4$, $\Gamma_5$, $\Gamma_6$, representing cases where each subspace in $\mathcal{S}$ overlaps with the subsequent subspace by 0, 1, 3, 5, 7, and 10 variables, respectively. For instance, $\Gamma_3=[3]\times19$. 
Based on table \ref{table:AOB}, the problem space $D$ is set to 1000 dimensions, with the variable search range being $\left[ -100,100 \right]$ in AOB. This avoids a reduction in the problem space dimension due to an increase in overlap.
$T_{\text{osz}}:\mathbb{R}^{D} \rightarrow \mathbb{R}^{D}$ is a transformation function to create smooth local irregularities \cite{hansen2009real} :
\begin{equation}
    \begin{aligned}
T_{\text {osz }}: x_{i} & \mapsto \operatorname{sign}\left(x_{i}\right) \exp \left(\hat{x}_{i}+0.049\left(\sin \left(c_{1} \hat{x}_{i}\right)+\sin \left(c_{2} \hat{x}_{i}\right)\right)\right)\\
\text { where } \hat{x}_{i} & =\left\{\begin{array}{ll}
\log \left(\left|x_{i}\right|\right) \text { if } x_{i} \neq 0\\
0 \text { otherwise }
\end{array}, \operatorname{sgin}(x)=\left\{\begin{array}{ll}
-1 & \text { if } x<0 \\
0 & \text { if } x=0 \\
1 & \text { if } x>0
\end{array}\right.\right. \\
c_{1} & =\left\{\begin{array}{ll}
10 & \text { if } x_{i}>0 \\
5.5 & \text { otherwise }
\end{array}, \text { and } c_{2}=\left\{\begin{array}{ll}
7.9 & \text { if } x_{i}>0 \\
3.1 & \text { otherwise } .
\end{array}\right.\right.
\end{aligned}
\end{equation}
$T_{\text{asy}}^{\beta}:\mathbb{R}^{D} \rightarrow \mathbb{R}^{D}$ is a transformation function to break the symmetry of the symmetric functions \cite{hansen2009real} :
\begin{equation}
    T_{\text {asy }}^{\beta}: x_{i} \mapsto\left\{\begin{array}{ll}
x_{i}^{1+\beta \frac{i-1}{D-1} \sqrt{x_{i}}} & \text { if } x_{i}>0 \\
x_{i} & \text { otherwise }
\end{array}\right.
\end{equation}
Different $\Gamma$ results in uneven subspace sizes, which in turn leads to different rotation matrices. Apart from $\Gamma$ and the corresponding generated $\mathcal{R}$, all other key components have only one set. Together, these key components, along with $\Gamma$ and $\mathcal{R}$, form 6 sets representing different levels of overlap. These 6 sets of key components are then combined with the 4 base functions to create 24 problems. For ease of subsequent description, we use the Schwefel function under the $\Gamma_3$ as an example, which we denote as $S_3$, i.e., the first letter of function name followed by $\Gamma_i$ with the index $i$.

\subsection{Experiment Setup and Analysis}
\subsubsection{Experiment Setup}
We conduct experiments on the 24 problems generated by AOB, as introduced in Section \ref{AOB}. We instantiate HCC as HCC-ES by using MM-ES as the NDA and CMAES as the OPT. The settings are exactly the same as those in Section \ref{section:anl of CEC}, and we compare HCC-ES with all previously mentioned algorithms. Table \ref{table_AOB} shows the performance of HCC-ES, with the best performance highlighted in bold. The "perf", "Time" and "Acc" represent the algorithm's mean performance and its standard deviation, running time, and decomposition accuracy, respectively. The symbols ``+'', ``-'', and ``$\approx$'' denote the outcomes of the Wilcoxon rank-sum test at the 0.05 significance level. The last column shows the test results for each algorithm, listing the number of times HCC-ES significantly outperformed competitors (+), instances with no significant difference ($\approx$), and cases where HCC-ES performed worse (-). 

\subsubsection{The Analysis of the Impact of Ideal Decomposition}
We first analyze and answer the first question in Section \ref{intro} based on table \ref{table_AOB}: is
ideal decomposition still important in this context?
We can draw the following outcomes:
 (1) Ideal decomposition is crucial. Firstly, the experimental results clearly demonstrate that RDDSM can achieve ideal decomposition. It is evident that DG2 fails in decomposing overlapping problems, severely affecting both its performance and time cost. Building on decomposition, the performance of the algorithm improves significantly from Random-CMAES to RDDSM-CMAES as the decomposition strategies and eventually achieves the ideal decomposition. It is consistent with the analysis of CEC2013LSGO in Section \ref{section:anl of CEC}.
(2) Incorrect decomposition has a significant impact on performance. Although decomposition is important, the performance comparison between DG2 and Random shows that random decomposition does not lead to performance improvement. The accuracy of decomposition is what truly matters. Another piece of evidence is the significant performance difference observed in RDG3-CMAES, despite the relatively small difference in decomposition between RDG3 and RDDSM when moving from near-ideal decomposition to achieving ideal decomposition.

\subsubsection{The Analysis of the Impact of Overlap}
We further analyze Table \ref{table_AOB} to address the remaining two questions in Section \ref{intro}. Clearly, HCC-ES outperforms all other algorithms significantly. We attribute this result to the following two factors: (1) For CC, decomposition is only the beginning in overlapping problems; how to perform cooperative optimization is equally important. This is because the coupling issue of the optimized values of overlapping variables significantly impacts the optimization performance. It can be observed that CC perform significantly better on non-overlapping problems than on overlapping ones, even when there is only one overlapping variable in each subspace (e.g., comparing S1 and S2). Even in problems like Ackley, where the optimization performance is relatively close, CC in A1 still shows a more improvement compared to other overlapping problems. Then, it can be observed that on all problems, as the degree of overlap increases, the objective value significantly increase, indicating a clear deterioration in algorithm performance. (2) For the overlapping problems, CC can overcome the existing optimization challenges through its combination with NDAs. This is essentially the result of balancing exploration and exploitation, which allows the full potential of NDAs and CC to be realized. It enables flexible adjustment based on $DO$, thus answering the question of when to use CC.
\begin{figure}
    \centering
    \includegraphics[width=1\linewidth]{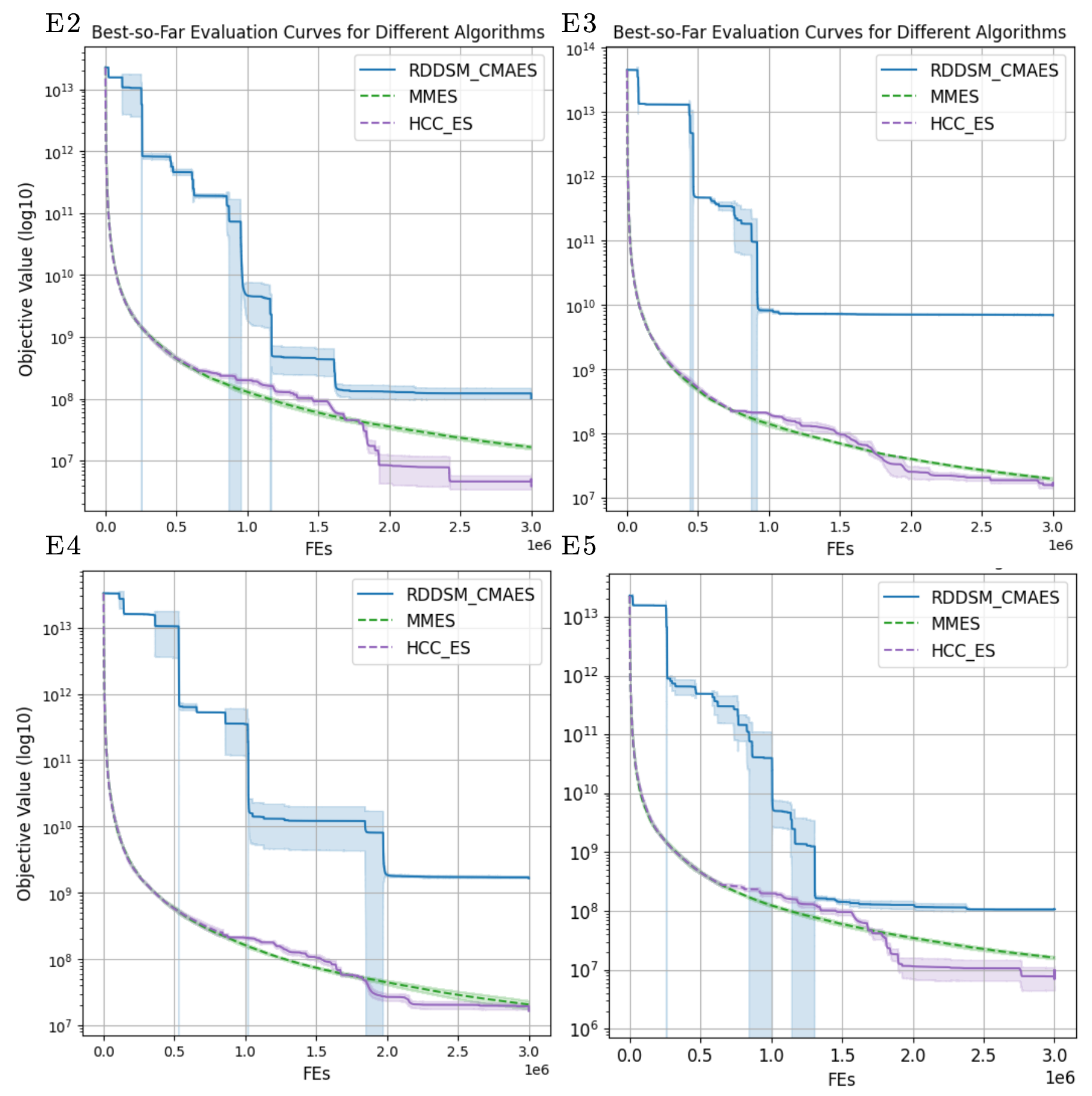}
    \caption{Comparison in Elliptic}
    \label{E2}
\end{figure}
Figure $\ref{E2}$ further illustrates the optimization characteristics of HCC in Elliptic as an example. Firstly, compared to CC (using RDDSM-CMAES as an example), HCC exhibits smaller variance. This may be attributed to the weighted sum of the optimization values, which results in smoother transitions of the coupling-optimized values of overlapping variables. Secondly, it fully demonstrates the advantages of HCC.
In the early stages, it can decrease significantly (dashed line), similar to NDAs. This leverages the exploitation capability of NDAs, avoiding the inefficient exploration in the early stages that consumes a large number of FEs in CC. However, when NDAs stagnate, the strong exploratory capability of CC allows HCC-ES to discover better solutions (solid line), resulting in the optimization performance surpassing that of either CC or NDAs alone.
\section{Conclusion and Future work}
\label{section:con}
We propose HCC, a novel two-phase cooperative co-evolution framework for large-scale global optimization with complex overlapping. It achieves stable ideal decomposition and, based on this, leverages decomposition information to mitigate the coupling issue of the optimized values of overlapping variables, while fully exploiting the development capability of NDAs and the exploration capability of CC. To analyze the impact of overlap and validate the performance of HCC, we extended the existing benchmark. Extensive experiments demonstrate that the algorithm instantiated within HCC significantly outperforms existing algorithms. The results reveal the characteristics of overlapping problems and highlight the distinct strengths of CC and NDAs. 

Looking ahead to future work, we hope to: (1) Further in-depth analysis and exploration of methods to address the coupling issue of the optimized values of overlapping variables. (2) Exploring more intelligent and diverse methods for balancing exploration and exploitation within HCC.

\begin{acks}
This work was supported in part by the National Natural Science
Foundation of China No. 62276100, in part by the Guangdong Provincial
Natural Science Foundation for Outstanding Youth Team Project No.
2024B1515040010, in part by the Guangdong Natural Science Funds for
Distinguished Young Scholars No. 2022B1515020049, and in part by the
TCL Young Scholars Program.
\end{acks}

\bibliographystyle{ACM-Reference-Format}
\bibliography{paper}

\end{document}